\documentclass[10pt,twocolumn,letterpaper]{article}

\usepackage[pagenumbers]{cvpr} 

\usepackage{graphicx}
\usepackage{amsmath}
\usepackage{amssymb}
\usepackage{booktabs}

\usepackage[linesnumbered,ruled,vlined]{algorithm2e}
\SetKwInput{KwInput}{Input}
\SetKwInput{KwOutput}{Output}
\usepackage{commath}

\usepackage{caption}
\usepackage{subcaption}
\usepackage{verbatim}
\usepackage{gensymb}
\usepackage{rotating}
\usepackage{marvosym}
\usepackage{multirow}
\usepackage{bm}
\usepackage{enumitem}

\usepackage[pagebackref,breaklinks,colorlinks]{hyperref}

\usepackage[capitalize]{cleveref}
\crefname{section}{Sec.}{Secs.}
\Crefname{section}{Section}{Sections}
\Crefname{table}{Table}{Tables}
\crefname{table}{Tab.}{Tabs.}

\def\cvprPaperID{11626} 
\def\confName{CVPR}
\def\confYear{2023}

\begin{document}
\title{Implicit 3D Human Mesh Recovery using Consistency with\\Pose and Shape from Unseen-view}

\author{Hanbyel Cho ~~~~~~~~~~~~~ Yooshin Cho ~~~~~~~~~~~~~ Jaesung Ahn ~~~~~~~~~~~~~ Junmo Kim\\
        Korea Advanced Institute of Science and Technology (KAIST), South Korea\\
        {\tt\small \texttt{\{\href{mailto:tlrl4658@kaist.ac.kr}{tlrl4658},\href{mailto:choys95@kaist.ac.kr}{choys95},\href{mailto:jaesung02@kaist.ac.kr}{jaesung02},\href{mailto:junmo.kim@kaist.ac.kr}{junmo.kim}\}@kaist.ac.kr}}
        }
\maketitle

\begin{abstract}
    From an image of a person, we can easily infer the natural 3D pose and shape of the person even if ambiguity exists. This is because we have a mental model that allows us to imagine a person's appearance at different viewing directions from a given image and utilize the consistency between them for inference. However, existing human mesh recovery methods only consider the direction in which the image was taken due to their structural limitations. Hence, we propose ``\textbf{Imp}licit 3D \textbf{H}uman \textbf{M}esh \textbf{R}ecovery (\textbf{ImpHMR})'' that can implicitly imagine a person in 3D space at the feature-level via Neural Feature Fields. In ImpHMR, feature fields are generated by CNN-based image encoder for a given image. Then, the 2D feature map is volume-rendered from the feature field for a given viewing direction, and the pose and shape parameters are regressed from the feature. To utilize consistency with pose and shape from unseen-view, if there are 3D labels, the model predicts results including the silhouette from an arbitrary direction and makes it equal to the rotated ground-truth. In the case of only 2D labels, we perform self-supervised learning through the constraint that the pose and shape parameters inferred from different directions should be the same. Extensive evaluations show the efficacy of the proposed method.
\end{abstract}
\vspace{-4mm}
\section{Introduction}
    Human Mesh Recovery (HMR) is a task that regresses the parameters of a three-dimensional (3D) human body model (\eg, SMPL~\cite{ref1_SMPL}, SMPL-X~\cite{ref_smplify-x}, and GHUM~\cite{ref_GHUM}) from RGB images. Along with 3D joint-based methods~\cite{ref_pavllo20193d,ref_CamDistHumanPose3D,ref_liu2020attention}, HMR has many downstream tasks such as AR/VR, and computer graphics as a fundamental topic in computer vision. In recent years, there has been rapid progress in HMR, particularly in regression-based approaches~\cite{ref3_HMR,ref4_SPIN,ref5_CMR,ref6_ProHMR,ref_HMR-ViT,ref_ROMP,ref_pymaf,ref_mpsnet,ref_hybrik,ref_ochmr}. However, despite these achievements, the existing algorithms still have a gap with the way humans do, so most of them do not show robust performance against the inherent ambiguity of the task.

\begin{figure}[!t]
    \centering
        \includegraphics[width=0.95\columnwidth]{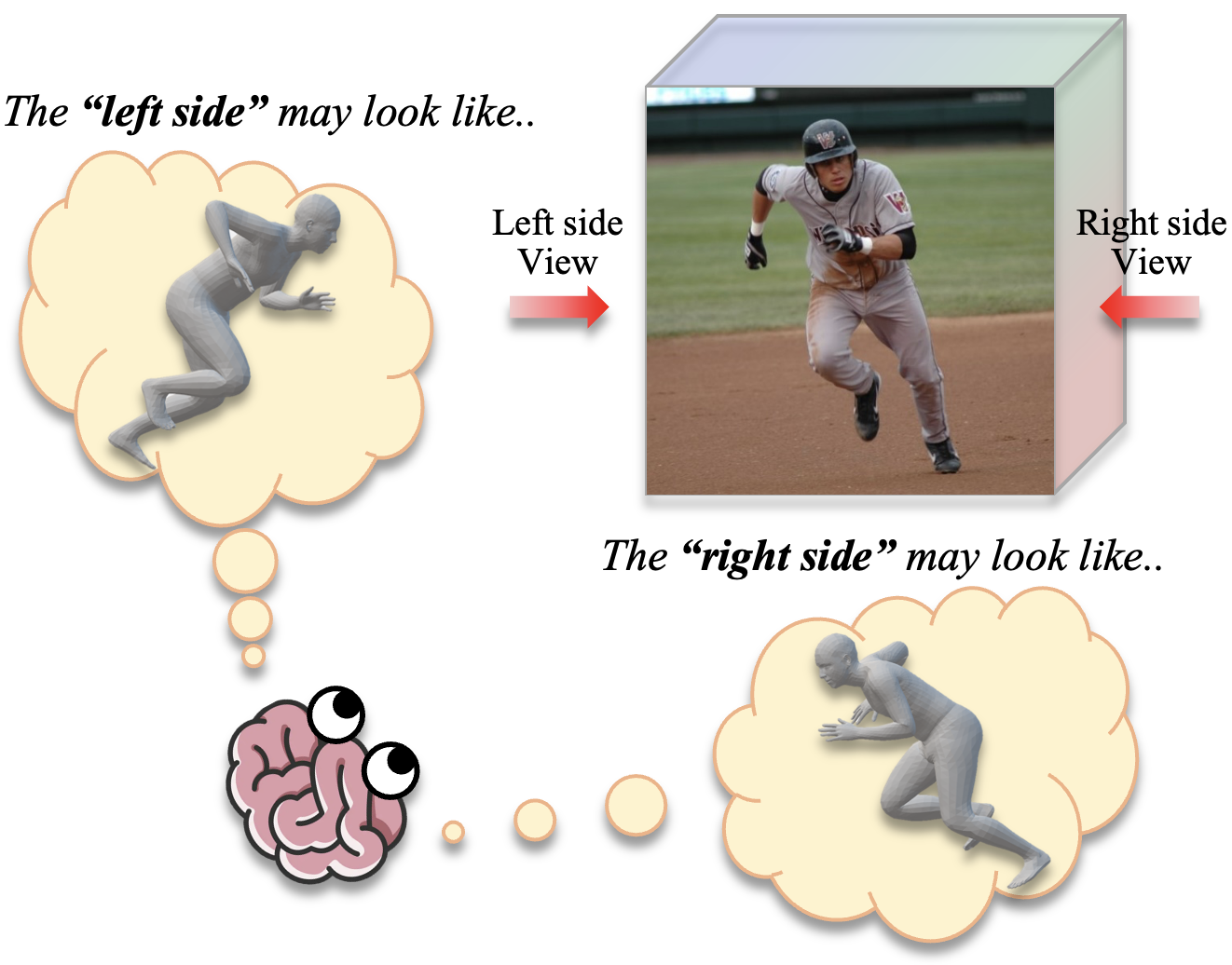}
        \vspace{-2mm}
        \caption{\textbf{Mental model of human that infers pose and shape from a single image.} From an image of a person, we infer pose and shape robustly by imagining the person’s appearance not only from the direction in which the image was taken, but also from other viewing directions (\eg, left and right sides).
        \vspace{-4mm}
        }
        \label{fig:teaser}
    \end{figure}

    Consider the image of a baseball player running, as shown in Fig.~\ref{fig:teaser}. For the given single image, we can easily infer that the person’s right elbow and left leg are extended backward in a 3D space, despite the presence of inherent ambiguity (\eg, depth and occlusion). This is because we have a mental model that allows us to imagine a person's appearance at different viewing directions from a given image and utilize the consistency between them for inference. Recently, many state-of-the-art studies have successfully utilized knowledge similar to that used by humans such as human dynamics~\cite{ref10_HMMR} and temporal information~\cite{ref7_VIBE,ref8_MEVA,ref9_TCMR,ref_mpsnet}. However, to the best of our knowledge, there have been no studies proposed methods that consider 3D space for HMR similar to the way we infer pose and shape through appearance check between different views in 3D space.

    To overcome this issue, we propose ``Implicit 3D Human Mesh Recovery (ImpHMR)'' that can implicitly imagine a human placed in a 3D space via Neural Feature Fields~\cite{ref16_GIR}. Our assumption is that if the model is trained to infer a human's pose and shape at arbitrary viewing directions in a 3D space from a single image, then the model learns better spatial prior knowledge about human appearance; consequently, the performance in the canonical viewing direction in which the image was taken is improved.
    
\begin{figure*}[!t]
    \centering
        \includegraphics[width=0.98\linewidth]{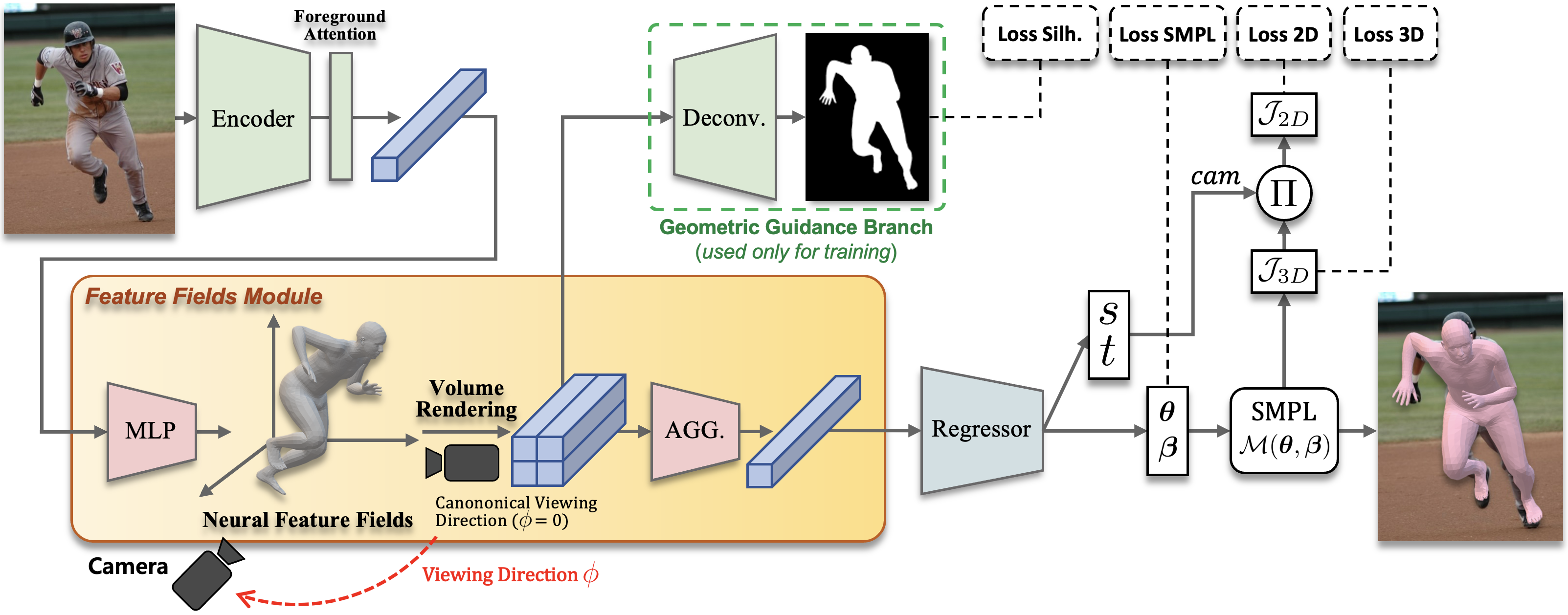}
        \vspace{-2mm}\caption{\textbf{Overview of ImpHMR architecture.}
        Given an image of a person, ImpHMR can implicitly imagine the person in 3D space and infer SMPL parameters viewed from an arbitrary viewing direction $\phi$ through \emph{Feature Fields Module}. The model infers parameters from arbitrary directions during training to have a better 3D prior about person; consequently, regression performance in \emph{Canonical Viewing Direction} is improved. For simplicity, we omit notation $\phi$ and write loss functions in Sec~\ref{sec:objective} abstractly according to the form of the output.
        }
        \vspace{-3mm}
        \label{fig:overall}
    \end{figure*}

    To achieve this, we incorporate Neural Feature Fields into regression-based HMR methods. In ImpHMR, it generates feature fields using a CNN-based image encoder for a given image to construct a person in 3D space, as shown in Fig.~\ref{fig:overall}. A feature field represented by a Multi-Layer Perceptron (MLP) is a continuous function that maps the position of a point in 3D space and a ray direction to a feature vector and volume density. In a feature field, which is implicit representation, all continuous points in a space can have a respective feature and volume density. Hence, the feature field is more expressive than explicit representation~\cite{ref_PTN} and more suitable for representing human appearance from different viewing directions in 3D space.

    To infer the pose and shape parameters from the Feature Field, the 2D feature map is generated by volume rendering for a given viewing direction, and the parameters are regressed from the rendered feature. Unlike previous methods, our model can look at a person from an \emph{arbitrary viewing direction} by controlling the viewing direction determined by camera extrinsic (\ie, camera pose). Therefore, to utilize consistency with pose and shape from unseen-view, if there are 3D labels, ImpHMR predicts results including silhouette used as \emph{geometric guidance} from an arbitrary direction and makes it equal to the rotated ground-truth. In addition, in the case of only 2D labels, we perform self-supervised learning through the constraint that SMPL parameters inferred from different directions should be the same. These constraints help feature fields represent a better 3D space by disentangling human appearance and viewing direction; as a result, SMPL regression from canonical viewing direction in which the image was taken is improved. To verify the efficacy of our method, we conduct experiments on {\small 3DPW, LSP, COCO}, and {\small 3DPW-OCC}. The contributions of our work can be summarized as follows:

    \vspace{-1mm}
    \begin{itemize}[leftmargin=*]
    \itemsep0em
        \vspace{-1mm}
        \item
        We propose a novel HMR model called ``ImpHMR'' that can implicitly imagine a human in 3D space from a given 2D observation via Neural Feature Fields.
        \vspace{-1mm}
        \item
        To utilize consistency with pose and shape from unseen-view, we propose arbitrary view imagination loss and appearance consistency loss.
        \vspace{-1mm}
        \item
        We propose the geometric guidance branch so that the model can learn better geometric information.
        \vspace{-1mm}
        \item
        ImpHMR has \emph{$2\sim3$ times faster} fps than current SOTAs thanks to efficient spatial representation in feature fields.
        \vspace{-1mm}
        \item
        We confirm that having the model imagine a person in 3D space and checking consistency between human appearance from different viewing directions improves the HMR performance in the canonical viewing direction in which the image was taken.
    \end{itemize}

\vspace{-2mm}
\section{Related Work}
    \subsection{Human Mesh Recovery}
        \vspace{-1mm}
        Human mesh recovery works have been conducted based on two approaches: \emph{optimization-based} approaches~\cite{ref2_SMPLify,rw_ref2_unite} and \emph{regression-based} approaches~\cite{ref3_HMR,ref15_NeuralBodyFit,ref13_LearningToEstim}. Recent works tend to focus on regression-based approaches.
        
        \vspace{-3mm}
        \paragraph{Optimization-based Approaches.} Early works in this field have mainly focused on the optimization-based approaches fitting parametric human body models. SMPLify~\cite{ref2_SMPLify} fits the parametric model, SMPL~\cite{ref1_SMPL} to minimize errors between the projection of recovered meshes and 2D/3D evidence, such as silhouettes or keypoints. In addition, prior terms are adopted to penalize the unrealistic shape and pose. In subsequent studies, 2D/3D information was utilized in the fitting procedure, and optimization with more expressive models in the multi-view has been suggested~\cite{rw_ref1_ADAM,rw_ref2_unite,rw_ref4_lightweight,rw_ref5_monocular}. Recently, hybrid approach, which combines optimization and regression-based approaches, has been proposed and provides a more accurate pseudo-ground-truth 3D (\eg, SPIN~\cite{ref4_SPIN} and EFT~\cite{ref_eft}) for 2D images. Despite the accurate results generated via optimization-based approaches, the fitting processes still remained slow and sensitive to initialization.
        
        \vspace{-3mm}
        \paragraph{Regression-based Approaches.} To avoid the issues of optimization-based methods, recent works have adopted regression-based approaches and utilized the powerful learning capability of deep neural networks~\cite{ref3_HMR,ref4_SPIN,ref15_NeuralBodyFit,ref13_LearningToEstim,ref12_Coherent,ref11_pose2mesh,ref14_MonoExp,ref_HMR-ViT}. Deep networks were directly used to regress model parameters from a single RGB image and supervised with 2D/3D annotations, such as 3D shape ground truth, keypoints, silhouettes, and parts segmentation. Regression-based methods have made significant advances by adopting network architectures that were suitable to learn different types of supervision signals~\cite{ref13_LearningToEstim,ref_coarse2fine,ref_delving,rw_ref8_BodyNet,ref_weakly,rw_ref6_I2l,rw_ref10_Denserac,ref_appearance,rw_ref9_Self,ref_texturepose}. Zhang~\etal\cite{ref_pymaf} proposed a pyramidal mesh alignment feedback that allows images and meshes to be well aligned, paying attention to the fact that there is no forward feedback when conventional regressors infer SMPL parameters iteratively. In addition, Li~\etal\cite{ref_hybrik} proposed a hybrid approach with joint estimation using inverse kinematics, and \cite{ref_ochuman,ref_pare} and \cite{ref_ochmr,ref_ROMP} proposed a method for a situation with occlusion and multi-person, respectively. Furthermore, recent studies have successfully utilized knowledge similar to that used by humans, such as human dynamics~\cite{ref10_HMMR} and temporal information~\cite{ref7_VIBE,ref8_MEVA,ref9_TCMR,ref_mpsnet}. However, to the best of our knowledge, there have been no studies that proposed methods that consider 3D space for HMR similar to the way we infer pose and shape through appearance checks between different views in 3D space.

    \subsection{Implicit Neural Representations}
        \paragraph{Neural Radiance Fields.} Previously, in 3D reconstruction, differentiable rendering techniques have been adopted to overcome the requirement for 3D supervision~\cite{rw_ref14_Diff_Rend,rw_ref15_Soft_Rasterizer}. A radiance field is a continuous function whose input is a set of a 3D location and a 2D ray direction, and its output is an RGB color value and a volume density~\cite{rw_ref13_OccupancyNetwork,rw_ref12_LearningImplicit}. To exploit the effective non-linear mapping capability of deep neural networks, Mildenhall~\etal~\cite{ref17_NeRF} proposed to learn Neural Radiance Fields (NeRFs) by parameterizing with a Multi-Layer Perceptron (MLP) and successfully combining with volume rendering for novel view synthesis.
        
        \vspace{-3mm}
        \paragraph{Neural Feature Fields.} Since the success of NeRFs~\cite{ref17_NeRF}, generative models for neural radiance fields have been proposed~\cite{rw_ref11_GRAF,ref16_GIR}. To get better representations of objects, Niemeyer~\etal\cite{ref16_GIR} proposed Generative Neural Feature Fields (GIRAFFE) that replace the color output with a generic feature vector. In addition, neural feature fields condition the MLP on latent vectors of the shape and appearance of objects. Therefore, unlike NeRFs that fits the MLP to multi-view images of a single scene, neural feature fields have the capability to generate novel scenes. In this study, we adopt Neural Feature Fields to design a mental model that imagines a human in a 3D space from a single image.
\vspace{-1.5mm}
\section{Methodology}
\vspace{-1.5mm}
The overall framework of the proposed method is shown in Fig.~\ref{fig:overall}. In this section, we provide a detailed explanation of the proposed method. First, we recapitulate the outline of Neural Feature Fields~\cite{ref16_GIR} and SMPL body model~\cite{ref1_SMPL}.
Then, we describe the model architecture and training objective of the proposed method.

\subsection{Neural Feature Fields}
A Neural Feature Field~\cite{ref16_GIR} is a continuous function 
$h$ that maps a 3D point $\mathbf{x} \in \mathbb{R}^3$ and a ray direction $\mathbf{r} \in \mathbb{S}^2$ to a volume density $\sigma \in \mathbb{R}^+$ and an $M_f$-dimensional feature vector $\mathbf{f} \in \mathbb{R}^{M_f}$. When $h$ is parameterized by a deep neural network, the low-dimensional input $\mathbf{x}$ and $\mathbf{r}$ are first mapped to higher-dimensional features through the positional encoding~\cite{ref17_NeRF,ref18_Fourier} so that they can be mapped to feature vectors $\mathbf{f}$ capable of representing complex scenes. Concretely, each element of $\mathbf{x}$ and $\mathbf{r}$ is mapped to a high-dimensional vector through the positional encoding, as follows:
\begin{align}
  \begin{split}
  \label{eq:pos-enc}
    \gamma(t, L) &= \\ (\sin(2^0 t \pi), &\cos(2^0 t \pi), \dots, \sin(2^L t \pi), \cos(2^L t \pi))
  \end{split}
\end{align}
where the scalar value $t$ is an element of $\mathbf{x}$ and $\mathbf{r}$, and $L$ is the number of frequency octaves.

Unlike Neural Radiance Fields (NeRFs)~\cite{ref17_NeRF} that outputs an RGB color value, Neural Feature Fields have the potential to be utilized in various downstream tasks because it outputs a feature vector for a given 3D point and a ray direction. For the generative perspective, Niemeyer~\etal\cite{ref16_GIR} proposed a novel generative model for Neural Feature Fields. In their model, called GIRAFFE, the object representations are represented by Neural Feature Fields (denoted as $h$) parameterized by Multi-Layer Perceptron (MLP). In order to express different objects (in our case, \emph{people with different poses and shapes}), the MLP is conditioned on latent vectors representing the object’s shape ($\mathbf{z}_s$) and appearance ($\mathbf{z}_a$) as follows:
\begin{align}\begin{split}
    h: \mathbb{R}^{L_\mathbf{x}} \times \mathbb{R}^{L_\mathbf{r}} \times \mathbb{R}^{M_s} \times \mathbb{R}^{M_a} &\to \mathbb{R}^+ \times \mathbb{R}^{M_f} \\ (\gamma(\mathbf{x}), \gamma(\mathbf{r}), \mathbf{z}_s, \mathbf{z}_a) &\mapsto (\sigma, \mathbf{f})
\end{split}\end{align}
where $L_{\mathbf{x}}$ and $L_{\mathbf{r}}$ denote output dimensions of the positional encodings; $M_s$ and $M_a$ denote dimension of $\mathbf{z}_s$ and $\mathbf{z}_a$ respectively; $\sigma$ and $\mathbf{f}$ denote a volume density and a feature vector. Finally, the model generates a realistic and controllable image through volume and neural rendering from inferred feature fields representing a specific scene.

In this work, we incorporate such a representation into conventional regression-based human mesh recovery so that the algorithm can implicitly imagine a person placed in a three-dimensional space.

\vspace{1mm}
\subsection{SMPL Body Model}
SMPL~\cite{ref1_SMPL} is a parametric human body model. It provides a function $\mathcal{M}(\boldsymbol{\theta}, \boldsymbol{\beta})$ that takes pose and shape parameters (denoted as $\boldsymbol{\theta} \in \mathbb{R}^{72}$ and $\boldsymbol{\beta} \in \mathbb{R}^{10}$ respectively) as inputs and outputs a body mesh $M \in \mathbb{R}^{6890 \times 3}$. The pose parameters consist of a global body rotation and 23 relative joint rotations. The shape parameters are the first 10 coefficients in the PCA shape space. For a given mesh, 3D joints $J$ can be obtained by linear combination of the mesh vertices, $J= WM$, with pre-trained linear regressor $W$.

\vspace{1mm}
\subsection{Model Architecture}
The intuition behind our method is that if the model is trained to infer a human’s pose and shape at arbitrary camera viewing directions in 3D space, the model learns spatial prior knowledge about human's appearance; consequently, the performance in the \emph{canonical viewing direction} in which the image was taken is improved. To achieve this, as depicted in Fig.~\ref{fig:overall}, our model mostly follows the HMR paradigm~\cite{ref3_HMR} where the input is an image of a person and output is the set of SMPL body model parameters, but there is a major difference in the \emph{Feature Fields Module}. In this section, we first describe the operation of the Feature Fields Module consisting of \emph{Feature Fields Generation} and \emph{Volume Rendering}, and then explain \emph{Parameter Regression} and \emph{Geometric Guidance Branch} using the module.

\vspace{-2mm}
\paragraph{Feature Fields Generation.} The goal of the Feature Fields Module is to construct the person in 3D space at the feature-level so that the model can look at the person from an arbitrary viewing direction. Given an input image $I$, we first encode the image using the CNN Encoder $g$ (\ie, ResNet-50~\cite{ref_resnet} \emph{before} the global average pooling) and obtain the feature vector $\mathbf{z} = g(I) \in \mathbb{R}^{2048 \times 7 \times 7}$. The encoded feature vector $\mathbf{z}$ may contain both information about the foreground and background of the given image. Thus, we use the \emph{Foreground Attention}~\cite{ref_CBAM} $\mathcal{A}$ and obtain the human-related feature vector $\mathbf{z}_{\text{fg}} = GAP(\mathcal{A}(\mathbf{z})) \in \mathbb{R}^{2048}$, where $GAP(\cdot)$ denotes global average pooling. Finally, the MLP (denoted as $h$) representing feature fields is conditioned on the latent vector $\mathbf{z}_{\text{fg}}$, and implicitly expresses the human in a 3D space.

\vspace{-1mm}
\paragraph{Volume Rendering.} In the feature field, we can look at the person represented by the feature field from an arbitrary viewing direction by controlling the camera pose (\ie, camera extrinsic). For the camera pose, since the ambiguity of the human pose occurs in the horizontal direction, we fix the elevation of the camera pose to $0$\textdegree\ and control only the azimuth (denoted as $\phi$). For simplicity, we denote the camera pose as $\phi$. Also, we define the direction in which the image was taken as \emph{Canonical Viewing Direction} (\small$\phi=0$\normalsize).

    \begin{figure}[!t]
    \centering
        \includegraphics[width=0.92\columnwidth]{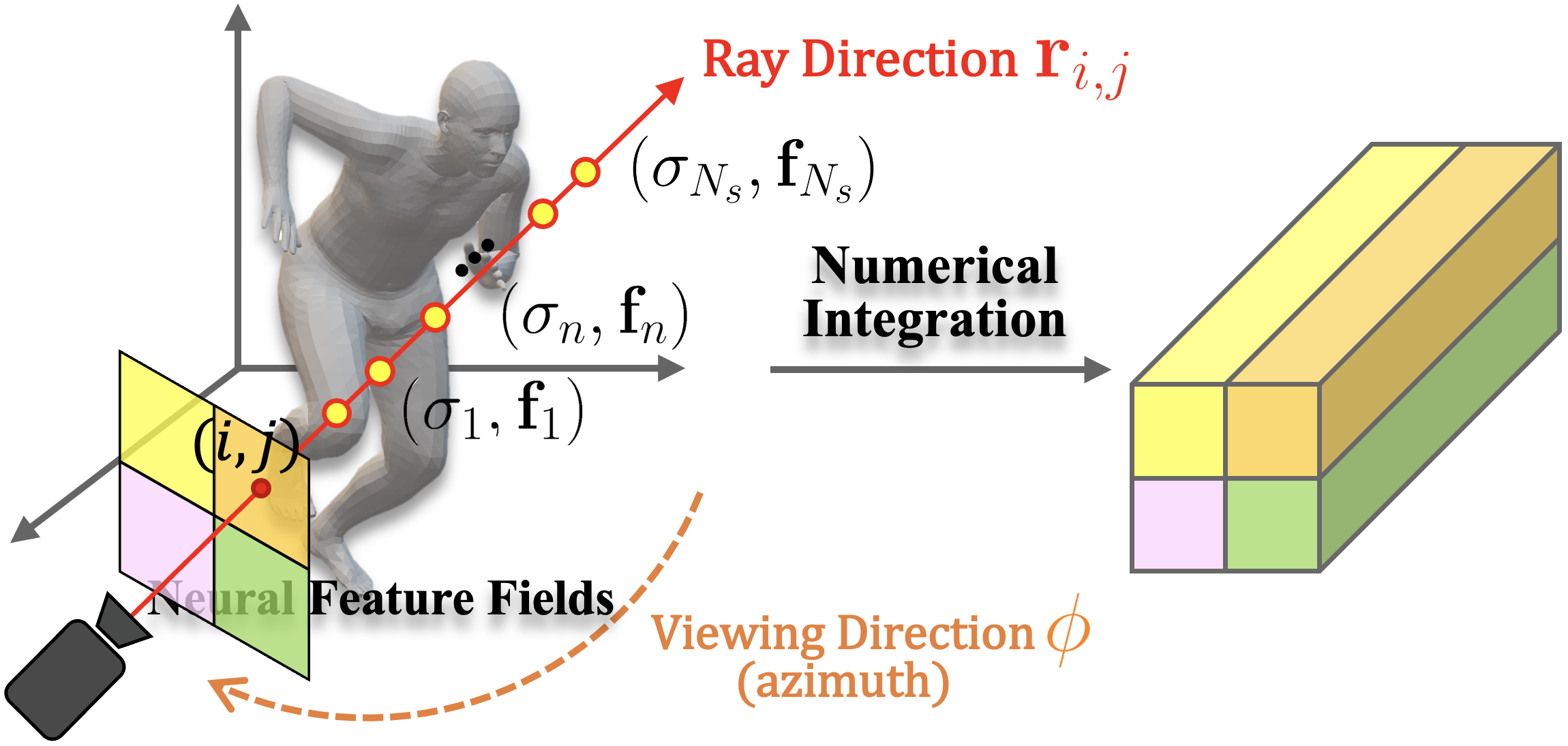}
        \vspace{-2mm}
        \caption{\textbf{Volume rendering procedure in a neural feature field.}
        To extract a 2D feature map from the Feature Field, sample points on the ray direction $\mathbf{r}_{i,j}$. From the volume density $\sigma$ and feature vector $\mathbf{f}$, the 2D feature map is obtained by Numerical Integration.
        }
        \label{fig:sample_points}
    \end{figure}

To infer the human pose and shape viewed from a viewing direction $\phi$, the 2D feature map $\mathbf{f}_{\phi} \in \mathbb{R}^{2048 \times H \times W}$ should be obtained from the feature field by volume rendering~\cite{ref17_NeRF}, where $H$ and $W$ denote the spatial resolutions of the feature. Given the camera pose $\phi$, let $\{\mathbf{x}_{i,j,n}\}_{n=1}^{N_s}$ be sample points on the ray direction $\mathbf{r}_{i,j}$ for the $(i,j)$ location of the 2D feature map, where $N_s$ is the number of sample points. We omit $i$ and $j$ for simplicity. Then, as shown in Fig.~\ref{fig:sample_points}, we can obtain a feature vector $\mathbf{f}_{n} \in \mathbb{R}^{2048}$ and a volume density $\sigma_{n}$ for each 3D point $\mathbf{x}$ as follows:
    \begin{align}
        \begin{split}\label{eq:sample_points}
            (\sigma_{n}, \mathbf{f}_{n}) = h(\gamma(\mathbf{x}_{n}), \gamma(\mathbf{r}), \mathbf{z}_{\text{fg}}).
        \end{split}
    \end{align}
where $\gamma$ denotes the positional encoding.

Finally, using \emph{Numerical Integration} as in \cite{ref16_GIR}, volume rendered feature vector $\mathbf{f}_{\text{rend}} \in \mathbb{R}^{2048}$ is obtained as follows:
    \begin{align}
        \small
        \begin{split}\label{eq:render-final-c}
            \mathbf{f}_{\text{rend}} = \sum_{n=1}^{N_s} \tau_n \alpha_n \mathbf{f}_n \quad
            \tau_n = \prod_{k=1}^{n-1} (1 - \alpha_k) \quad \alpha_n = 1 - e^{-\sigma_n\delta_n}
        \end{split}
        \normalsize
    \end{align}
where $\tau_n$ is the transmittance, $\alpha_n$ the alpha value for $\mathbf{x}_n$, and $\delta_n = {\left|\left|\mathbf{x}_{n+1} - \mathbf{x}_n \right|\right|}_2$ the distance between neighboring sample points. We repeat this process for each spatial location $(i,j)$ and obtain the 2D feature map $\mathbf{f}_{\phi} \in \mathbb{R}^{2048 \times H \times W}$, as in~\cite{ref16_GIR}. Finally, to use the feature map for SMPL parameter regression, we generate feature vector $\mathbf{z}_{\phi} = AGG(\mathbf{f}_{\phi}) \in \mathbb{R}^{2048}$, where $AGG(\cdot)$ denotes aggregation layer consisting of single depthwise convolution.

\begin{figure}[!t]
    \centering
        \includegraphics[width=0.98\columnwidth]{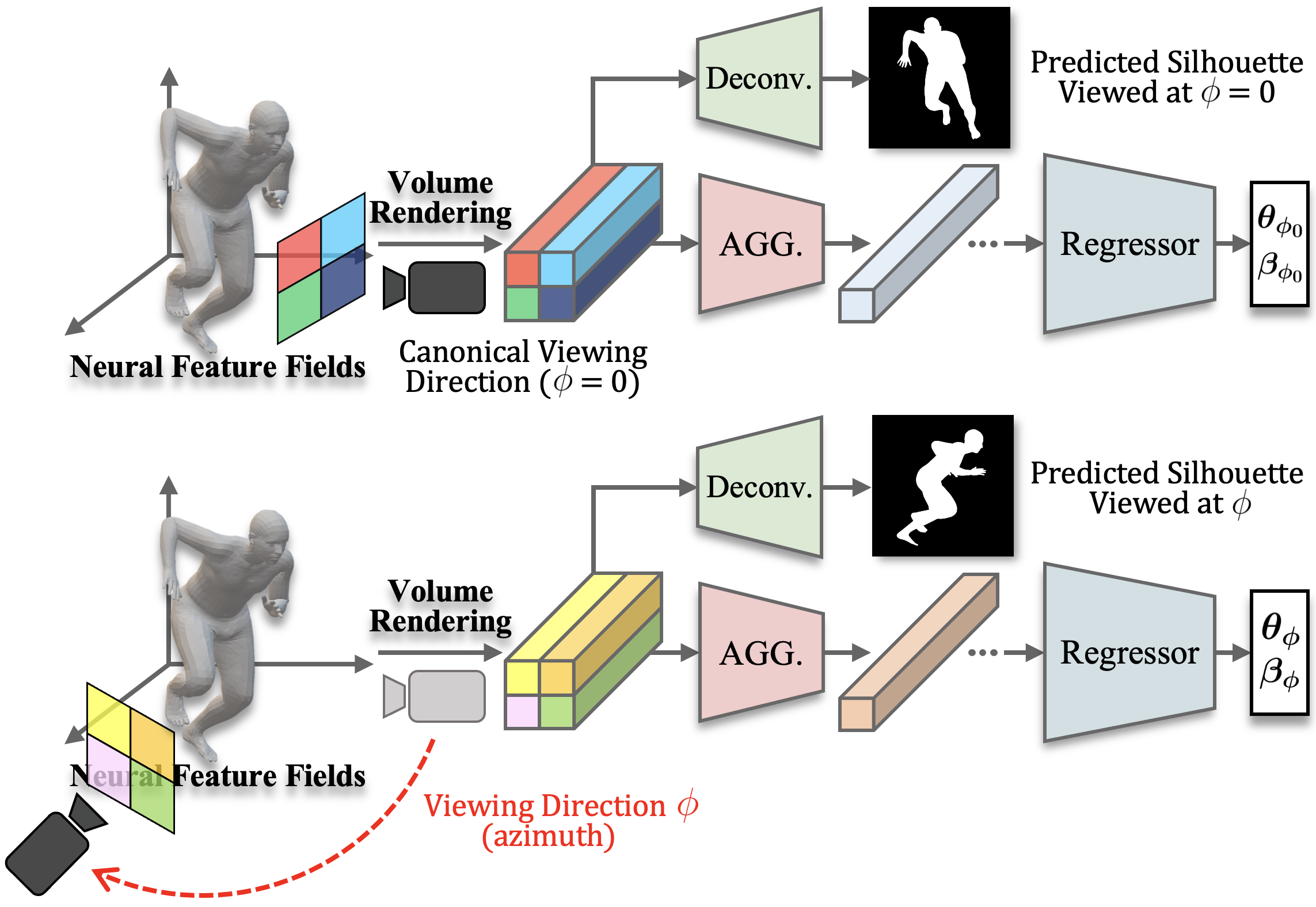}
        \vspace{-2mm}
        \caption{\textbf{SMPL parameter and silhouette regression with controlling camera viewing direction.}
        \textbf{Top:} regression from the \emph{Canonical Viewing Direction} ($\phi=0$), as in conventional methods. \textbf{Bottom:} regression from an arbitrary viewing direction.
        }
        \vspace{-4mm}
        \label{fig:param_reg}
    \end{figure}

\vspace{-1mm}
\paragraph{Parameter Regression.} We have obtained the feature vector $\mathbf{z}_{\phi}$ that contains the information about the person viewed from the camera pose $\phi$ through \emph{Feature Fields Module}. From the feature vector $\mathbf{z}_{\phi}$, ImpHMR predicts the SMPL model $\Theta_\phi=\{\bm{\theta}_\phi, \bm{\beta}_\phi, \bm{\pi}_\phi\}$ using the regressor $\mathcal{R}(\cdot)$ as $\Theta_\phi = \mathcal{R}(\mathbf{z}_{\phi})$, where each element of $\Theta_\phi$ denotes the pose, shape, and camera parameters inferred from the viewing direction $\phi$ respectively. Note that, when \small$\phi=0$\normalsize\ (in short, \small$\phi_0$\normalsize), ImpHMR outputs $\Theta_{\phi_0}$ that is the inference result viewed from the direction in which the image was taken, as in conventional regression-based methods, and otherwise outputs the inference result viewed from a viewing direction $\phi$, as shown in Fig.~\ref{fig:param_reg}. Therefore, after the training, we use the $\phi$ fixed to $0$ for testing.

From the predicted parameters $\Theta$, we can generate the body mesh with vertices $M=\mathcal{M}(\bm{\theta}, \bm{\beta}) \in \mathbb{R}^{6890 \times 3}$. Subsequently, using the pre-trained linear regressor, 3D joints $J \in \mathbb{R}^{N_j\times 3}$ ($\mathcal{J}_{3D}$ in Fig.~\ref{fig:overall}) can be regressed from the mesh vertices $M$, where $N_j$ is the number of joints. Furthermore, 2D keypoints $K \in \mathbb{R}^{N_j\times 2}$ ($\mathcal{J}_{2D}$ in Fig.~\ref{fig:overall}) are obtained as $K =\bm{\Pi}(J)$, where $\bm{\Pi}(\cdot)$ denotes the projection function from weak-perspective camera parameters $\bm{\pi} \in [s,t]$ ($s$ and $t$ denote the scale and translation parameters, respectively).

\begin{figure}[!t]
    \centering
        \includegraphics[width=\columnwidth]{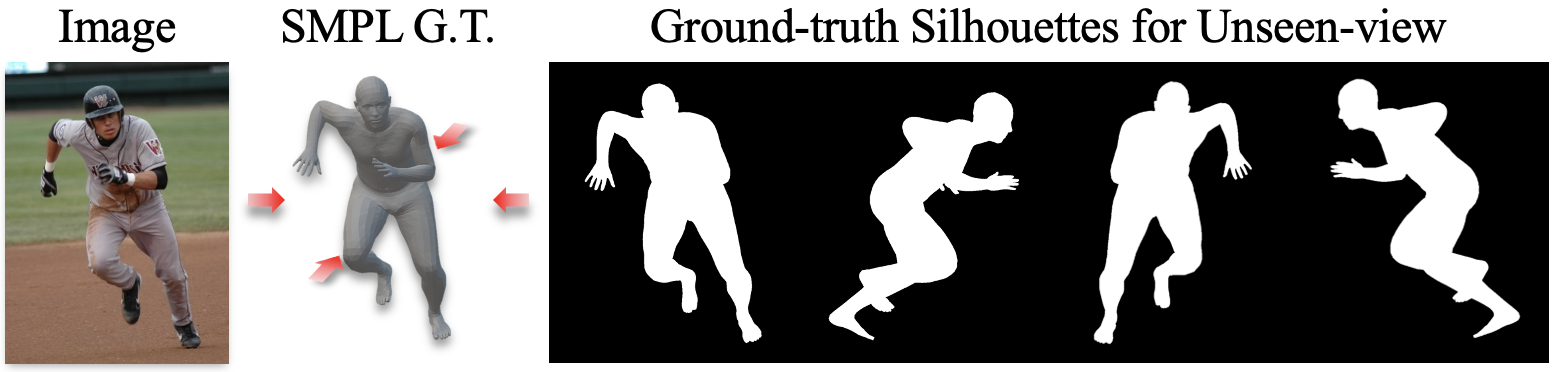}
        \vspace{-6mm}
        \caption{\textbf{Generating ground-truth silhouettes viewed from unseen-view.}
        G.T. silhouette from an arbitrary viewing direction is generated by rotating the mesh of SMPL G.T. and rendering it.
        }
        \vspace{-0mm}
        \label{fig:silh_gt}
    \end{figure}

\vspace{-4mm}
\paragraph{Geometric Guidance Branch.} ImpHMR is trained to regress the rotated ground-truth viewed from an arbitrary viewing direction $\phi$ to learn spatial prior of human's appearance (see Sec.~\ref{sec:objective}). However, unlike GIRAFFE, which generates images, our model regresses parameters (\ie, SMPL), so there might not be enough information for the model to learn the geometry of 3D space. Thus, as shown in Fig.~\ref{fig:param_reg}, we have the model reconstruct the silhouette $S_\phi$ (viewed at direction $\phi$) from the 2D feature map $\mathbf{f}_\phi$ using deconvolution $\mathcal{D}(\cdot)$ as $S_\phi = \mathcal{D}(\mathbf{f}_\phi)$. To explicitly give geometric supervision for \emph{unseen-view}, we generate the G.T. silhouette from the G.T. SMPL mesh rotated by the viewing direction using NMR~\cite{ref_renderer}, as shown in Fig.~\ref{fig:silh_gt}. Note that, the geometric guidance branch is used \emph{only} for training.

\subsection{Training Objective}
\label{sec:objective}
The final goal of our method is to improve the regression performance in the \emph{canonical viewing direction} ($\phi_0$) by having the model learn spatial prior about the person in 3D space. In this section, we describe the following three objectives for training: \emph{Canonical View Regression}, \emph{Arbitrary View Imagination}, and \emph{Appearance Consistency Loss}.

\vspace{-1mm}
\paragraph{Canonical View Regression Loss.} This is the constraint for inference from \emph{canonical viewing direction} ($\phi_0$) just like previous methods~\cite{ref3_HMR,ref4_SPIN}. 2D keypoints $K_{\phi_0}$ and 3D joints $J_{\phi_0}$ are obtained from inferred SMPL parameters (\ie, \small$\bm{\theta}_{\phi_0}$\normalsize, \small$\bm{\beta}_{\phi_0}$\normalsize, and \small$\bm{\pi}_{\phi_0}$\normalsize), making them close to their G.T. as follows:
    \begin{equation}
        \begin{split}
            \mathcal{L}_{reg} = \lambda_{2d}||K_{\phi_0} - \hat{K}|| + \lambda_{3d}||J_{\phi_0} - \hat{J}||\quad\quad\quad\\
            \quad\quad\quad\quad+ \lambda_{pose}||\bm{\theta}_{\phi_0} - \hat{\bm{\theta}}|| + \lambda_{shape}||\bm{\beta}_{\phi_0} - \hat{\bm{\beta}}||,
        \label{eq:loss_reg}
        \end{split}
    \end{equation}
where $||\cdot||$ is the squared L2 norm; $\hat{K}$, $\hat{J}$, $\hat{\bm{\theta}}$, and $\hat{\bm{\beta}}$ denote the ground-truth 2D keypoints, 3D joints, and SMPL pose and shapes, respectively following the notation of \cite{ref_pymaf}.

\vspace{-1mm}
\paragraph{Arbitrary View Imagination Loss.} To leverage the consistency of pose and shape from unseen-views, we train the model to infer human’s appearance viewed at arbitrary directions. Thus, if \emph{3D labels exist}, we use the constraint that the predicted result from an arbitrary viewing direction $\phi$ (sampled from the distribution of camera pose $p_{cam}$) should be equal to the ground-truth rotated by $\small-\normalsize\phi$ as follows:
    \begin{equation}
        \begin{split}
            \mathcal{L}_{imag} = \mathbb{E}_{\phi \sim p_{cam}}\lbrack \lambda_{3d}||J_{\phi} - \hat{J}_{\scriptscriptstyle-\scriptstyle\phi}|| + \lambda_{silh.}||S_{\phi} - \hat{S}_{\scriptscriptstyle-\scriptstyle\phi}||\\
            + \lambda_{pose}||\bm{\theta}_{\phi} - \hat{\bm{\theta}}_{\scriptscriptstyle-\scriptstyle\phi}|| + \lambda_{shape}||\bm{\beta}_{\phi} - \hat{\bm{\beta}}|| \rbrack,\quad\quad\quad
        \label{eq:loss_dis}
        \end{split}
    \end{equation}
where $\hat{J}_{\scriptscriptstyle-\scriptstyle\phi}$ is ground-truth 3D joints rotated by $-\phi$ in horizontal direction, $\hat{S}_{\scriptscriptstyle-\scriptstyle\phi}$ is G.T. silhouette viewed at $-\phi$, and $\hat{\bm{\theta}}_{\scriptscriptstyle-\scriptstyle\phi}$ is ground-truth pose rotated by $-\phi$ only for the global orientation, and \small$p_{cam} \sim \mathcal{U}[0,2\pi]$\normalsize. Using the constraint, we can disentangle human appearance and viewing direction, resulting in better spatial prior about humans in 3D space.

Note that we rotate the ground-truth by $-\phi$ because the viewing direction and shown person's appearance are rotated oppositely. In addition, for the ground-truth shape parameters $\hat{\bm{\beta}}$, we do not apply the rotation because it is independent of the viewing direction.

\vspace{-1mm}
\paragraph{Appearance Consistency Loss.}
ImpHMR can make predictions in various viewing directions from a single image. By utilizing this capability of the model, we perform \emph{self-supervised} learning through the constraint that SMPL parameters inferred from different directions (sampled from $p_{cam}$) should be the same if there are \emph{only 2D labels} as:
    \begin{equation}
        \begin{split}
            \mathcal{L}_{cons} = \mathbb{E}_{\phi_1,\phi_2 \sim p_{cam}}\lbrack \lambda_{pose}||\bm{\theta}'_{\phi_1} - \bm{\theta}_{\phi_2}||
            \quad\quad\quad\\
            + \lambda_{shape}||\bm{\beta}_{\phi_1} - \bm{\beta}_{\phi_2}|| \rbrack,
        \label{eq:loss_ss}
        \end{split}
    \end{equation}
where $\bm{\theta}'_{\phi_1}$ denotes the modified parameters where only the global orientation of the inferred pose parameters $\bm{\theta}_{\phi_1}$ is changed by \small$\phi_2-\phi_1$\normalsize\ amount.
Finally, our overall loss function is \small$\mathcal{L}_{all} = \mathcal{L}_{reg} + \mathcal{L}_{imag} + \mathcal{L}_{cons}$\normalsize. We selectively use each loss function depending on whether 3D labels are available or not and our model is trained \emph{end-to-end} manner.
\section{Experiments}
    \subsection{Datasets and Evaluation Metrics}
        \label{exp:datasets and evaluation}
        Following previous works~\cite{ref3_HMR,ref4_SPIN,ref_pare}, we use a mixture of 2D and 3D datasets. We use MPI-INF-3DHP~\cite{ref_mpii3d} and Human3.6M~\cite{ref_h36m} with ground-truth SMPL as our 3D datasets for training. Also, MPII~\cite{ref_mpii2d}, COCO~\cite{ref_COCO}, and LSPET~\cite{ref_lspet} with the pseudo-ground-truth SMPL provided by \cite{ref_eft} are used as 2D datasets. As in PARE~\cite{ref_pare}, we divide the training process into two phases to reduce the overall training time. We first train our model on COCO for ablation studies, and then obtain the final performance using a mixture of all datasets for comparison with SOTA methods. For evaluation, we use 3DPW~\cite{ref_3dpw} and 3DPW-OCC~\cite{ref_ochuman} for quantitative evaluation. Our method is evaluated using mean per joint position error (MPJPE), Procrustes-aligned mean per joint position error (PA-MPJPE), and per-vertex error (PVE) metrics. For qualitative evaluation, we evaluate the quality of the inferred mesh on 3DPW, LSP~\cite{ref_LSP}, and COCO validation sets. More description about datasets is in the supplementary material.

    \subsection{Experimental Results}
        \label{sec:experiment results}

        \begin{table}[!t]
        \centering
            \resizebox{0.96\columnwidth}{!}{
            \begin{tabular}{cl|r|r|r}
            \toprule
			& & \multicolumn{3}{c}{ \small 3PDW }  \\
		\cmidrule(lr){3-5}
			&\small \bf Method &{\small MPJPE $\downarrow$} & {\small PA-MPJPE $\downarrow$} & {\small PVE $\downarrow$} \\
            \midrule
                \multirow{7}[2]{*}{\begin{sideways}\rotatebox[origin=c]{0}{Temporal}\end{sideways}}
                    & HMMR~\cite{ref10_HMMR}      & 116.5 & 72.6 & 139.3  \\
                    & DSD~\cite{ref_dsd}          & -     & 69.5 & -      \\
                    & Arnab~\etal~\cite{ref_arnab}& -     & 72.2 & -      \\
                    & Doersch~\etal~\cite{ref_doersch}& - & 74.7 & -      \\
                    & VIBE~\cite{ref7_VIBE}       & 93.5  & 56.5 & 113.4  \\
                    & TCMR~\cite{ref9_TCMR}       & 95.0  & 55.8 & 111.3  \\
                    & MPS-Net~\cite{ref_mpsnet}   & 91.6  & 54.0 & 109.6  \\
            \midrule
                \multirow{10}[4]{*}{\begin{sideways}\rotatebox[origin=c]{0}{Frame-based}\end{sideways}}
                    & HMR~\cite{ref3_HMR}         & 130.0 & 76.7 & -      \\
                    & GraphCMR~\cite{ref5_CMR}    & -     & 70.2 & -      \\
                    & SPIN~\cite{ref4_SPIN}       & 96.9  & 59.2 & 116.4  \\
                    & PyMAF~\cite{ref_pymaf}      & 92.8  & 58.9 & 110.1  \\
                    & I2L-MeshNet~\cite{rw_ref6_I2l} & 100.0   & 60.0 & - \\
                    & ROMP~\cite{ref_ROMP}        & 89.3 & 53.5 & 105.6   \\
                    & HMR-EFT~\cite{ref_eft}      & -    & 54.2 & -       \\
                    & PARE~\cite{ref_pare}  & \underline{82.9} & \underline{52.3} & \underline{99.7} \\
            \cmidrule{2-5}
                    & ImpHMR (Ours)           & \textbf{81.8} & \textbf{49.8} & \textbf{96.4} \\
                    & ImpHMR (Ours) w. 3DPW    & 74.3 & 45.4 & 87.1 \\
            \bottomrule
            \end{tabular}
            }
            \vspace{-0.05in}
            \caption{\textbf{Results on 3DPW.}
            Best in bold, second-best underlined. Values are in mm. {\footnotesize ``ImpHMR (Ours)''} and {\footnotesize ``ImpHMR (Ours) w. 3DPW''} denote the model trained \emph{w/o} and \emph{w.} 3DPW train set, respectively.
            \vspace{-6mm}
            }
            \vspace{-1mm}
            \label{tab:sota_quant_3dpw}
        \end{table}

        \begin{figure}[!t]
        \centering
            \includegraphics[width=\columnwidth]{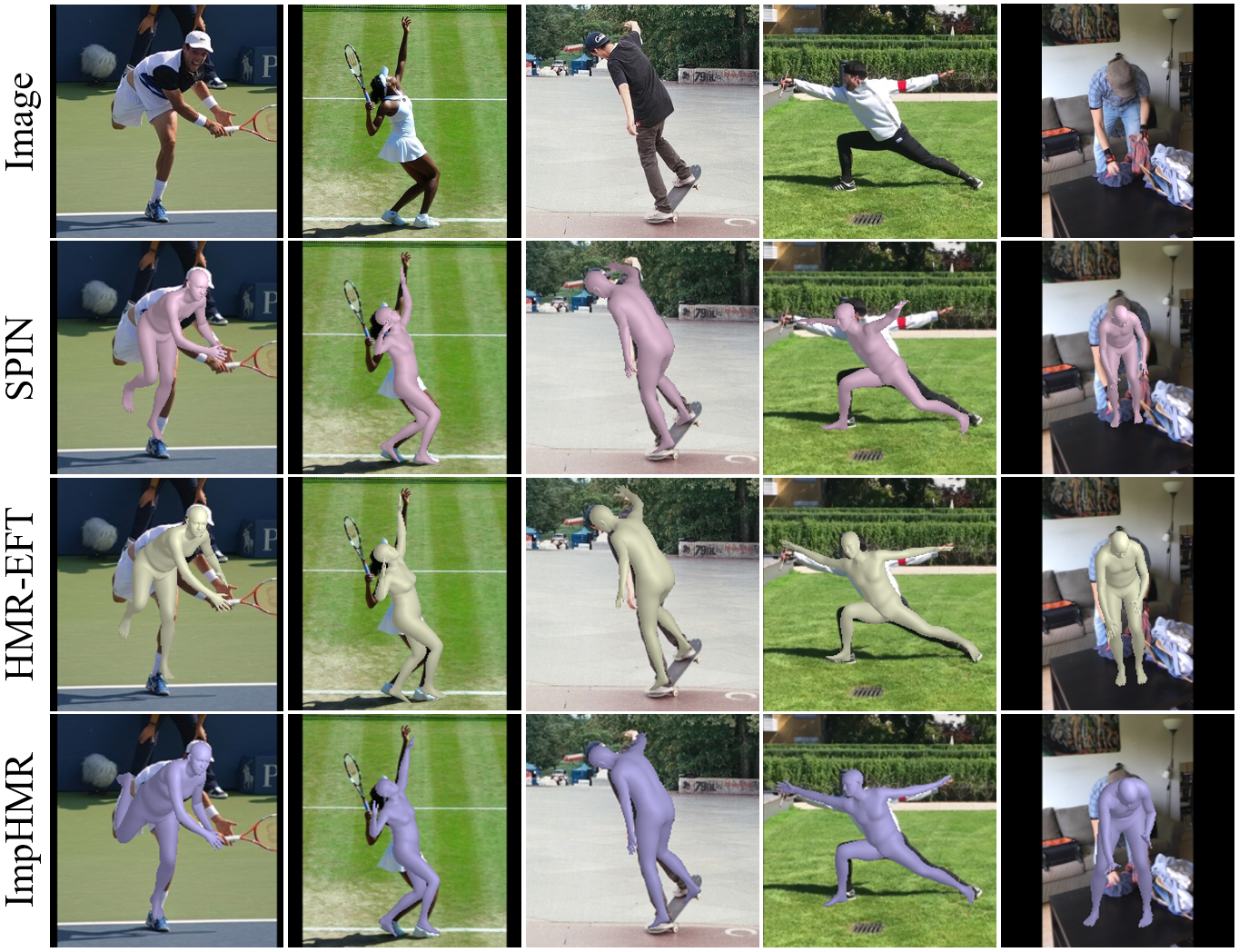}
            \vspace{-6mm}
            \caption{\textbf{Qualitative results.}
            Qualitative comparison of the proposed method with SPIN~\cite{ref4_SPIN} and HMR-EFT~\cite{ref_eft} on COCO validation set and 3DPW test split.
            \vspace{-1mm}
            }
            \vspace{-1mm}
            \label{fig:qualitative}
        \end{figure}

            In this section, we validate the effectiveness of the proposed method. First, we compare the performance of ImpHMR with previous SOTA methods. Then, we confirm whether ImpHMR has the ability to infer human appearance viewed from different viewing directions in 3D space. Finally, the efficacy of each of the methods is validated.

        \begin{figure}[!t]
        \centering
            \includegraphics[width=.97\columnwidth]{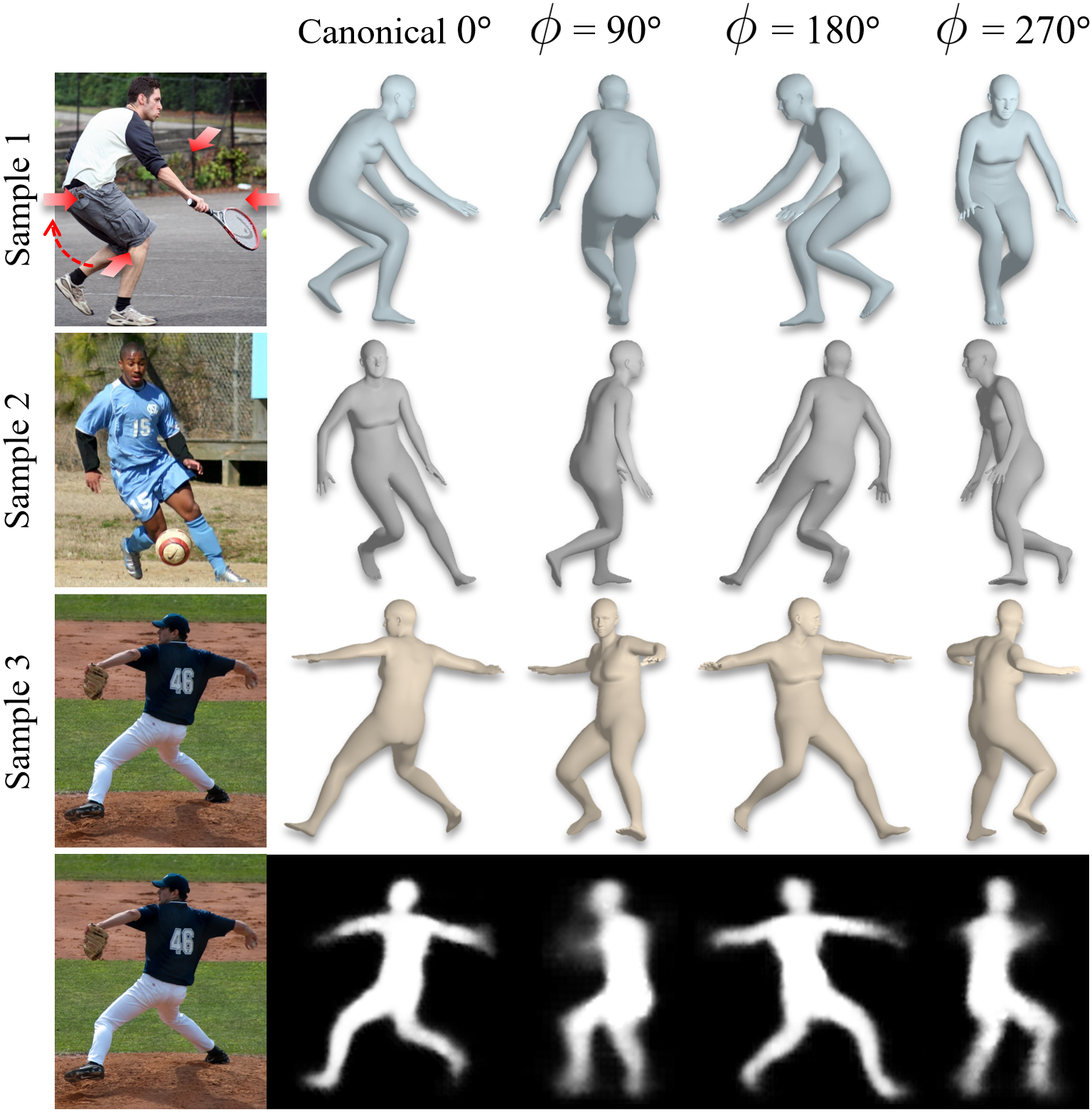}
            \vspace{-1.5mm}
            \caption{\textbf{Inferred SMPL mesh and  silhouettes viewed from different viewing directions.}
            Results inferred \emph{by changing the viewing direction} clockwise by $90^\circ$ from canonical viewing direction. Note that the inference results are \emph{not by rotating the mesh} inferred from the canonical viewing direction, but \emph{directly inferring a person viewed from different directions in 3D space}.
            }
            \vspace{-4mm}
            \label{fig:imagination}
        \end{figure}

        \vspace{-3mm}
        \paragraph{Comparison with State-of-the-Art.}
            First, we evaluate the human mesh recovery performance of ImpHMR. Table~\ref{tab:sota_quant_3dpw} shows the quantitative results of previous state-of-the-art and our method on 3DPW test split. As shown in Tab.~\ref{tab:sota_quant_3dpw}, our method (denoted as {\small ``ImpHMR (Ours)''}) shows superior performance compared to other methods for all metrics in both temporal- and frame-based approaches. In particular, ImpHMR shows an -4.4mm (8.1\%) performance improvement in PA-MPJPE metric compared to HMR-EFT~\cite{ref_eft}. Also, it shows a -1.1mm (1.3\%), -2.5mm (4.8\%), and -3.3mm (3.3\%) improvement in MPJPE, PA-MPJPE, and PVE, respectively, compared to PARE~\cite{ref_pare}, the most previous best performing model. Also, we report the results of the model trained using 3DPW train split (denoted as {\small ``ImpHMR (Ours) w. 3DPW''} in Tab.~\ref{tab:sota_quant_3dpw}) to see the performance of the model when using the ground-truth SMPL labels. Compared to when the dataset is not used, ImpHMR shows a significant performance improvement and outperforms all methods by a large margin.

            We perform an evaluation on 3DPW-OCC~\cite{ref_ochuman}, an occlusion-specific dataset, to verify the performance of ImpHMR in the presence of ambiguity (\eg, occlusion). Table~\ref{tab:sota_qaunt_3dpw_occ} shows the result. For fair comparison, all methods in the table are trained on the same datasets (\ie, Human3.6M~\cite{ref_h36m}, COCO~\cite{ref_COCO}, and, 3DOH~\cite{ref_ochuman}). As shown in Tab.~\ref{tab:sota_qaunt_3dpw_occ}, ImpHMR outperforms the occlusion-specific methods Zhang~\etal~\cite{ref_ochuman} and PARE~\cite{ref_pare}, including HMR-EFT~\cite{ref_eft}, by a large margin. This demonstrates that the structure and learning method of ImpHMR is suitable for modeling situations in which ambiguity is present.

            For qualitative comparisons, we compare our method with SPIN~\cite{ref4_SPIN} and HMR-EFT~\cite{ref_eft}. As shown in the Fig.~\ref{fig:qualitative}, ImpHMR outputs a mesh that is well aligned with the image even when a person with extreme poses or ambiguity exists.

        \vspace{-3.5mm}
        \paragraph{Results from Different Viewing Directions.}
            In this section, we verify that ImpHMR successfully imagines a person in 3D space. To do this, we report the mesh reconstruction results from different viewing directions $\phi$ (\ie, $0^\circ$, $90^\circ$, $180^\circ$, and $270^\circ$) for a given image of a person. As shown in Fig.~\ref{fig:imagination}, we confirm that ImpHMR can imagine a person's appearance not only from the canonical viewing direction, but also from the image from the left, right, and back of the person. Note that the inference results are \emph{not by rotating the mesh} inferred from the canonical viewing direction, but the results of \emph{directly inferring a person viewed from different directions in 3D space}. A viewing direction can be an arbitrary angle; herein, we report only the results from the $4$ different angles.

            Additionally, the last row of Fig.~\ref{fig:imagination} is the person's silhouettes in Sample $3$ viewed from various directions by using $\mathcal{D}(\cdot)$, and it can be seen that the inferred silhouettes are similar to what a human imagines. This demonstrates that spatially meaningful information is contained in a volume-rendered 2D feature map $\mathbf{f}_\phi$ by an appropriate guide of Geometric Guidance Branch.

        \begin{table}[!t]
        \centering
            \resizebox{0.8\columnwidth}{!}{
            \begin{tabular}{@{}l|r|r|r@{}}
                \toprule
			      ~~~\small \bf Method & {\small MPJPE $\downarrow$} & {\small PA-MPJPE $\downarrow$} & {\small PVE $\downarrow$} ~~\\
			\midrule
			      ~~~ Zhang~\etal~\cite{ref_ochuman} & -     & 72.2  & -     ~~\\
                    ~~~ HMR-EFT~\cite{ref_eft}         & 94.4  & 60.9  & 111.3 ~~\\
                    ~~~ PARE~\cite{ref_pare}     & 90.5  & 56.6  & 107.9 ~~\\
			\midrule
                    ~~~ ImpHMR (Ours)                   & \textbf{86.5} & \textbf{54.4} & \textbf{104.7} ~~\\
			\bottomrule
	       \end{tabular}
            }
            \vspace{-0.05in}
            \caption{\textbf{Results on 3DPW-OCC.}
            For fair comparison, all methods are trained using the same datasets (\ie, Human3.6M, COCO, and 3DOH). Best in bold.
            }
            \label{tab:sota_qaunt_3dpw_occ}
            \vspace{-1.5mm}
        \end{table}{}

        \begin{table}[!t]
        \centering
            \tabcolsep=1mm
            \resizebox{0.95\columnwidth}{!}{
            \begin{tabular}{@{}l|r|r|r|r@{}}
            \toprule
                ~~~\small \bf Method ~&~ {\small SPL. $1$} ~~&~ {\small SPL. $2$} ~~&~ {\small SPL. $3$} ~~&~ {\small LSP dataset} ~~\\
            \midrule
                ~~~$\mathcal{L}_{reg}$ ~&~ 0.0763 ~&~ 0.0875 ~&~ 0.0711 ~&~ 0.0844 ~~\\
                ~~~$\mathcal{L}_{reg} + \mathcal{L}_{imag}$ ~&~ 0.0618 ~&~ 0.0512 ~&~ 0.0639 ~&~ 0.0774 ~~\\
                ~~~$\mathcal{L}_{reg} + \mathcal{L}_{imag} + \mathcal{L}_{cons}$ ~~&~ 0.0361 ~&~ 0.0314 ~&~ 0.0445 ~&~ 0.0561 ~~\\
            \bottomrule
            \end{tabular}
            }
            \vspace{-0.05in}
            \caption{\textbf{Entanglement between shape and viewing direction.}
            Each value denotes the degree of \emph{variation of the inferred shape} when inferred by changing the viewing direction. All methods are trained using COCO dataset.
            }
            \vspace{-1.5mm}
            \label{tbl:entanglement}
        \end{table}

        \begin{table}[!t]
        \centering
            \resizebox{0.95\columnwidth}{!}{
            \begin{tabular}{@{}cl|r|r|r@{}}
            \toprule
                &\small \bf Method &{\small MPJPE $\downarrow$} & {\small PA-MPJPE $\downarrow$} & {\small PVE $\downarrow$} ~~\\
            \midrule
                \multirow{1}[1]{*}{\begin{sideways}\rotatebox[origin=c]{0}{}\end{sideways}}
                 & \small Baseline                    & \small 101.6 & \small 58.3 & \small 117.2 ~~\\
            \midrule
                \multirow{3}[1]{*}{\begin{sideways}\rotatebox[origin=c]{0}{\small\ w/o $\mathcal{D}(\cdot)$}\end{sideways}}
                 & $\mathcal{L}_{reg}$                           & \small 96.5 & \small 58.9 & \small 116.0 ~~\\
                 & $\mathcal{L}_{reg} + \mathcal{L}_{imag}$      & \small 94.9 & \small 57.9 & \small 114.2 ~~\\
                 & $\mathcal{L}_{reg} + \mathcal{L}_{imag} + \mathcal{L}_{cons}$     & \small 93.5 & \small 57.5 & \small 113.1 ~~\\
            \midrule
                \multirow{1}[1]{*}{\begin{sideways}\rotatebox[origin=c]{0}{}\end{sideways}}
                 & $\mathcal{L}_{reg} + \mathcal{L}_{imag} + \mathcal{L}_{cons}$     & \small 92.7 & \small 57.0 & \small 112.1 ~~\\
            \bottomrule
            \end{tabular}
            }
            \vspace{-0.05in}
            \caption{\textbf{Effectiveness of each proposed method.}
            The results are evaluated on the 3DPW dataset. Values for all metrics are in mm. w/o $\mathcal{D}(\cdot)$ denotes the method trained without Geometric Guidance Branch. All methods are trained using COCO dataset.
            }
            \vspace{-3mm}
            \label{tbl:contrib}
        \end{table}

            To quantitatively verify the 3D spatial construction capability of ImpHMR, we measure the \textbf{E}ntanglement between the \textbf{S}hape and the \textbf{V}iewing direction (in short, ESV). If the 3D space is well constructed in Neural Feature Fields, the body shape should be consistent despite changes in viewing direction. Therefore, we change the viewing direction from $0^\circ$ to $360^\circ$ at $1^\circ$ intervals and define the average of the standard deviations of the inferred shape parameters as ESV, which is the degree of entanglement. We measure the ESV for each sample image in Fig.~\ref{fig:imagination} and the \emph{entire} LSP~\cite{ref_LSP} dataset. As shown in Tab.~\ref{tbl:entanglement}, we can notice that the degree of entanglement decreases as the proposed constraints are added. This indicates that ImpHMR successfully disentangles body shape and viewing direction, as a result imagining a person in a 3D space well. A detailed description of ESV is in the supplementary material.

        \begin{figure}[!t]
        \centering
            \includegraphics[width=0.94\columnwidth]{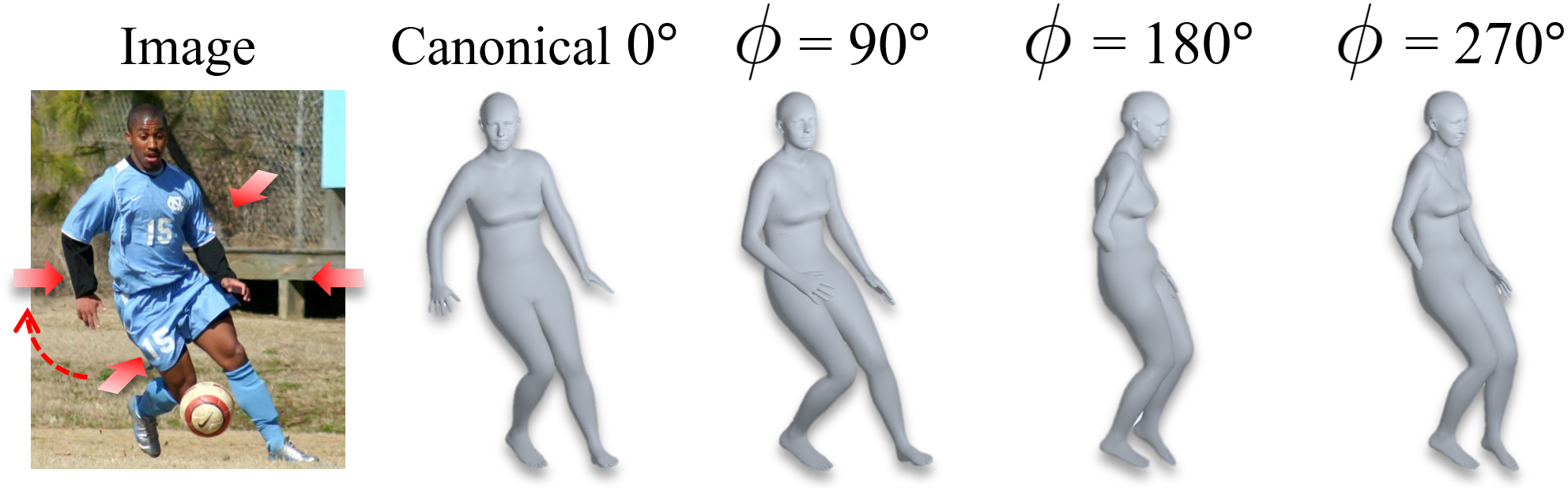}
            \vspace{-2mm}
            \caption{\textbf{Inferred SMPL mesh from different viewing directions with \emph{Explicit} representation.}
            Results inferred by changing the direction clockwise by $90^\circ$ from canonical viewing direction.
            }
            \vspace{-4.5mm}
            \label{fig:imagination_explicit}
        \end{figure}

        \vspace{-3mm}
        \paragraph{Ablation Studies.}
            To verify the efficacy of each of the proposed methods, we evaluate the performance change by adding each method. For a fair comparison, we train a baseline model (denoted as {\small ``Baseline''} in Tab.~\ref{tbl:contrib}) that has the same model architecture and the number of parameters as ImpHMR (except {\small $AGG(\cdot)$}) but does not perform feature fields generation and volume rendering. As shown in Tab.~\ref{tbl:contrib}, we can notice that all methods provide a positive contribution. Compared to Baseline, it can be seen that there is an improvement even when just generating a feature vector through volume rendering within feature fields (denoted as {\small $\mathcal{L}_{reg}$}). This verifies the inference method of ImpHMR is suitable for HMR tasks. In addition, as shown in Tab.~\ref{tbl:entanglement} and Tab.~\ref{tbl:contrib}, by adding the proposed constraints, including silhouette loss with \emph{geometric guidance branch}, the better the model disentangles the person’s appearance and viewing direction in 3D space, and the performance increases accordingly. Through this, we can confirm that our assumption about the proposed method is valid.

            Table~\ref{tab:ablation_model_arch} shows the ablation of the model architecture (\ie, aggregation layer {\small $AGG$} and foreground attention {\small $\mathcal{A}$}). We use three types of {\small $AGG$}: global average pooling (GAP), convolution (Conv.), and depth-wise convolution (DWConv.), and report the performance of combinations with {\small $\mathcal{A}$}. As can be seen in Tab.~\ref{tab:ablation_model_arch}, {\small $AGG$} shows good performance in the order of DWConv, GAP, and Conv. We can notice that foreground attention has a positive effect except for Conv. We finally adopted the best-performing set of Tab.~\ref{tab:ablation_model_arch} (e).

            ImpHMR uses neural feature fields, an implicit representation, to imagine a person in 3D. However, as a means of expressing 3D space, there is also an explicit representation such as the voxel-based method (\eg, PTN~\cite{ref_PTN}). To explore the suitability of explicit representation, we check the performance of the baseline in which Feature Fields Module of ImpHMR is replaced by a voxel-based representation. For the volumetric representation, we use Perspective Transformer Nets~\cite{ref_PTN} (PTN). For a fair comparison, we set the voxel resolution to $4\times4\times4$, the same as ImpHMR. Since PTN can perspective project features for a given camera extrinsic, we train the baseline using the same constraints (\ie, $\mathcal{L}_{reg}$, $\mathcal{L}_{imag}$, and $\mathcal{L}_{cons}$\normalsize) as in ImpHMR. Figure~\ref{fig:imagination_explicit} shows the inference results of the baseline, and we can notice that it fails to model 3D space. This indicates that implicit representation is more suitable for modeling a person in 3D.

        \begin{table}[!t]
        \centering
            \resizebox{0.95\columnwidth}{!}{
            \begin{tabular}{ll|r|r|r}
            \toprule
                \small \bf Method & & {\small MPJPE $\downarrow$} & {\small PA-MPJPE $\downarrow$} & {\small PVE $\downarrow$} \\
            \midrule
			& HMR-EFT~\cite{ref_eft}             & 99.0 & 59.9 & - ~~~\\
		  \midrule
                \bf $AGG$ Aggregation & \bf $\mathcal{A}$ Attention & \multicolumn{3}{c}{}  \\
		  \midrule
			(a) GAP      & with  $\mathcal{A}$   & 95.3 & \underline{57.6} & 114.5 \\
			(b) GAP      & without $\mathcal{A}$ & 95.8 & 57.9 & 114.9 \\
		  \midrule
			(c) Conv.    & with $\mathcal{A}$    & 96.0 & 58.0 & 114.6 \\
			(d) Conv.    & without $\mathcal{A}$ & 95.2 & 57.7 & 114.9 \\
		  \midrule
			(e) DWConv.  & with $\mathcal{A}$    & \textbf{92.7} & \textbf{57.0} & \textbf{112.1} \\
			(f) DWConv.  & without $\mathcal{A}$ & \underline{94.7} & 57.9 & \underline{113.8} \\
		  \bottomrule
            \end{tabular}
            }
            \vspace{-0.06in}
            \caption{\textbf{Ablation study of the model architecture.}
            ``$AGG$ Aggregation'' denotes the type of aggregation layer in volume rendering. ``$\mathcal{A}$ Attention'' denotes whether foreground attention is used. All methods are trained using COCO dataset.}
            \vspace{-1mm}
            \label{tab:ablation_model_arch}
        \end{table}{}

        \begin{table}[!t]
        \centering
            \resizebox{0.90\columnwidth}{!}{
            \begin{tabular}{@{}l|r|r|r|r@{}}
            \toprule
                ~~~\small \bf Method ~&~ {\small Res.$1$} ~~&~ {\small Res.$2$} ~~&~ {\small Res.$4$} ~~&~ {\small Res.$6$} ~~~\\
            \midrule
                ~~~HMR-EFT~\cite{ref_eft}   ~~&~ 115.5  ~&~ - ~&~ - ~&~ - ~~\\
                ~~~PyMAF~\cite{ref_pymaf}   ~~&~ 33.6   ~&~ - ~&~ - ~&~ - ~~\\
                ~~~PARE~\cite{ref_pare}     ~~&~ 27.5   ~&~ - ~&~ - ~&~ - ~~\\
            \midrule
                ~~~ImpHMR (Ours)             ~~&~ 88.8   ~&~ 88.4 ~&~ 87.1 ~&~ 78.2 ~~\\
            \bottomrule
            \end{tabular}
            }
            \vspace{-0.06in}
            \caption{\textbf{Comparison of inference speed.}
            The numbers are in \emph{frames per second} (\textbf{fps}). The Res. denotes the spatial resolution of a 2D feature map in volume rendering for our method. Thanks to efficient spatial representation in feature fields, ImpHMR shows about \emph{$2\sim3$ times faster} fps compared to PyMAF and PARE.
            }
            \vspace{-4mm}
            \label{tbl:inference_time}
        \end{table}

            Table~\ref{tbl:inference_time} compares the inference speed between ImpHMR and the current SOTA methods. For fair evaluation, frames per second (fps) is calculated by averaging the time it took for each model to infer $10000$ times of an input image of $224\times224$ size on RTX 2080Ti GPU. As shown in Tab.~\ref{tbl:inference_time}, we can notice that ImpHMR is slightly slower than HMR-EFT, but still has \emph{real-time} performance. Especially, ImpHMR has \emph{$2\sim3$ times faster} fps than PyMAF and PARE, which are current SOTA methods. This is because ImpHMR is capable of efficient spatial representation within neural feature fields compared to the latest SOTA methods that utilize spatial information.

\section{Conclusion and Future Works}
\vspace{-1mm}
We have introduced a novel HMR model called ``ImpHMR'' that can implicitly imagine a human in 3D space from a given 2D observation via neural feature fields. To utilize consistency with pose and shape from unseen-views, we propose arbitrary view imagination loss and appearance consistency loss. Also, we propose geometric guidance branch that helps the model can learn better geometric information. ImpHMR has $2\sim3$ times faster fps than current SOTAs thanks to efficient spatial representation in feature fields. Also, extensive evaluation proves that our method is valid. For future works, we can make a more occlusion-robust model by carefully modeling volume density.

\noindent{\textbf{Acknowledgements.} This work was conducted by Center for Applied Research in Artificial Intelligence (CARAI) grant funded by DAPA and ADD (UD190031RD).

\clearpage
{\small
\bibliographystyle{ieee_fullname}
\bibliography{egbib}

\begin{thebibliography}{10}\itemsep=-1pt

\bibitem{ref_mpii2d}
Mykhaylo Andriluka, Leonid Pishchulin, Peter Gehler, and Bernt Schiele.
\newblock 2d human pose estimation: New benchmark and state of the art
  analysis.
\newblock In {\em IEEE Conference on Computer Vision and Pattern Recognition
  (CVPR)}, June 2014.

\bibitem{ref_arnab}
Anurag Arnab, Carl Doersch, and Andrew Zisserman.
\newblock Exploiting temporal context for 3d human pose estimation in the wild.
\newblock In {\em Proceedings of the IEEE/CVF Conference on Computer Vision and
  Pattern Recognition (CVPR)}, June 2019.

\bibitem{ref2_SMPLify}
Federica Bogo, Angjoo Kanazawa, Christoph Lassner, Peter~V. Gehler, Javier
  Romero, and Michael~J. Black.
\newblock Keep it smpl: Automatic estimation of 3d human pose and shape from a
  single image.
\newblock In {\em ECCV}, 2016.

\bibitem{rw_ref14_Diff_Rend}
Wenzheng Chen, Huan Ling, Jun Gao, Edward Smith, Jaakko Lehtinen, Alec
  Jacobson, and Sanja Fidler.
\newblock Learning to predict 3d objects with an interpolation-based
  differentiable renderer.
\newblock {\em Advances in Neural Information Processing Systems},
  32:9609--9619, 2019.

\bibitem{rw_ref12_LearningImplicit}
Zhiqin Chen and Hao Zhang.
\newblock Learning implicit fields for generative shape modeling.
\newblock In {\em Proceedings of the IEEE/CVF Conference on Computer Vision and
  Pattern Recognition}, pages 5939--5948, 2019.

\bibitem{ref_HMR-ViT}
Hanbyel Cho, Jaesung Ahn, Yooshin Cho, and Junmo Kim.
\newblock Video inference for human mesh recovery with vision transformer.
\newblock In {\em 2023 IEEE 17th International Conference on Automatic Face and
  Gesture Recognition (FG)}, pages 1--6, 2023.

\bibitem{ref_CamDistHumanPose3D}
Hanbyel Cho, Yooshin Cho, Jaemyung Yu, and Junmo Kim.
\newblock Camera distortion-aware 3d human pose estimation in video with
  optimization-based meta-learning.
\newblock In {\em Proceedings of the IEEE/CVF International Conference on
  Computer Vision}, pages 11169--11178, 2021.

\bibitem{ref9_TCMR}
Hongsuk Choi, Gyeongsik Moon, Ju~Yong Chang, and Kyoung~Mu Lee.
\newblock Beyond static features for temporally consistent 3d human pose and
  shape from a video.
\newblock In {\em Proceedings of the IEEE/CVF Conference on Computer Vision and
  Pattern Recognition (CVPR)}, pages 1964--1973, June 2021.

\bibitem{ref11_pose2mesh}
Hongsuk Choi, Gyeongsik Moon, and Kyoung~Mu Lee.
\newblock Pose2mesh: Graph convolutional network for 3d human pose and mesh
  recovery from a 2d human pose.
\newblock In Andrea Vedaldi, Horst Bischof, Thomas Brox, and Jan-Michael Frahm,
  editors, {\em Computer Vision -- ECCV 2020}, pages 769--787, Cham, 2020.
  Springer International Publishing.

\bibitem{ref14_MonoExp}
Vasileios Choutas, Georgios Pavlakos, Timo Bolkart, Dimitrios Tzionas, and
  Michael~J. Black.
\newblock Monocular expressive body regression through body-driven attention.
\newblock In Andrea Vedaldi, Horst Bischof, Thomas Brox, and Jan-Michael Frahm,
  editors, {\em Computer Vision -- ECCV 2020}, pages 20--40, Cham, 2020.
  Springer International Publishing.

\bibitem{ref_doersch}
Carl Doersch and Andrew Zisserman.
\newblock Sim2real transfer learning for 3d human pose estimation: motion to
  the rescue.
\newblock In H. Wallach, H. Larochelle, A. Beygelzimer, F. d\textquotesingle
  Alch\'{e}-Buc, E. Fox, and R. Garnett, editors, {\em Advances in Neural
  Information Processing Systems}, volume~32. Curran Associates, Inc., 2019.

\bibitem{ref_resnet}
Kaiming He, Xiangyu Zhang, Shaoqing Ren, and Jian Sun.
\newblock Deep residual learning for image recognition.
\newblock {\em CoRR}, abs/1512.03385, 2015.

\bibitem{ref_bn}
Sergey Ioffe and Christian Szegedy.
\newblock Batch normalization: Accelerating deep network training by reducing
  internal covariate shift.
\newblock In Francis Bach and David Blei, editors, {\em Proceedings of the 32nd
  International Conference on Machine Learning}, volume~37 of {\em Proceedings
  of Machine Learning Research}, pages 448--456, Lille, France, 07--09 Jul
  2015. PMLR.

\bibitem{ref_h36m}
Catalin Ionescu, Dragos Papava, Vlad Olaru, and Cristian Sminchisescu.
\newblock Human3.6m: Large scale datasets and predictive methods for 3d human
  sensing in natural environments.
\newblock {\em IEEE Transactions on Pattern Analysis and Machine Intelligence},
  36(7):1325--1339, jul 2014.

\bibitem{ref12_Coherent}
Wen Jiang, Nikos Kolotouros, Georgios Pavlakos, Xiaowei Zhou, and Kostas
  Daniilidis.
\newblock Coherent reconstruction of multiple humans from a single image.
\newblock In {\em Proceedings of the IEEE/CVF Conference on Computer Vision and
  Pattern Recognition (CVPR)}, June 2020.

\bibitem{ref_LSP}
Sam Johnson and Mark Everingham.
\newblock Clustered pose and nonlinear appearance models for human pose
  estimation.
\newblock In {\em Proc. BMVC}, pages 12.1--11, 2010.
\newblock doi:10.5244/C.24.12.

\bibitem{ref_lspet}
Sam Johnson and Mark Everingham.
\newblock Learning effective human pose estimation from inaccurate annotation.
\newblock In {\em CVPR 2011}, pages 1465--1472, 2011.

\bibitem{ref_eft}
Hanbyul Joo, Natalia Neverova, and Andrea Vedaldi.
\newblock Exemplar fine-tuning for 3d human pose fitting towards in-the-wild 3d
  human pose estimation.
\newblock In {\em 3DV}, 2020.

\bibitem{rw_ref1_ADAM}
Hanbyul Joo, Tomas Simon, and Yaser Sheikh.
\newblock Total capture: A 3d deformation model for tracking faces, hands, and
  bodies.
\newblock In {\em Proceedings of the IEEE conference on computer vision and
  pattern recognition}, pages 8320--8329, 2018.

\bibitem{ref3_HMR}
Angjoo Kanazawa, Michael~J. Black, David~W. Jacobs, and Jitendra Malik.
\newblock End-to-end recovery of human shape and pose.
\newblock In {\em Computer Vision and Pattern Regognition (CVPR)}, 2018.

\bibitem{ref10_HMMR}
Angjoo Kanazawa, Jason~Y. Zhang, Panna Felsen, and Jitendra Malik.
\newblock Learning 3d human dynamics from video.
\newblock In {\em Proceedings of the IEEE/CVF Conference on Computer Vision and
  Pattern Recognition (CVPR)}, June 2019.

\bibitem{ref_renderer}
Hiroharu Kato, Yoshitaka Ushiku, and Tatsuya Harada.
\newblock Neural 3d mesh renderer.
\newblock In {\em The IEEE Conference on Computer Vision and Pattern
  Recognition (CVPR)}, 2018.

\bibitem{ref_ochmr}
Rawal Khirodkar, Shashank Tripathi, and Kris Kitani.
\newblock Occluded human mesh recovery.
\newblock In {\em Proceedings of the IEEE/CVF Conference on Computer Vision and
  Pattern Recognition (CVPR)}, pages 1715--1725, June 2022.

\bibitem{ref7_VIBE}
Muhammed Kocabas, Nikos Athanasiou, and Michael~J. Black.
\newblock Vibe: Video inference for human body pose and shape estimation.
\newblock In {\em Proceedings of the IEEE/CVF Conference on Computer Vision and
  Pattern Recognition (CVPR)}, June 2020.

\bibitem{ref_pare}
Muhammed Kocabas, Chun-Hao~P. Huang, Otmar Hilliges, and Michael~J. Black.
\newblock {PARE}: Part attention regressor for {3D} human body estimation.
\newblock In {\em Proc. International Conference on Computer Vision (ICCV)},
  pages 11127--11137, Oct. 2021.

\bibitem{ref4_SPIN}
Nikos Kolotouros, Georgios Pavlakos, Michael~J. Black, and Kostas Daniilidis.
\newblock Learning to reconstruct 3d human pose and shape via model-fitting in
  the loop.
\newblock In {\em Proceedings of the IEEE International Conference on Computer
  Vision}, 2019.

\bibitem{ref5_CMR}
Nikos Kolotouros, Georgios Pavlakos, and Kostas Daniilidis.
\newblock Convolutional mesh regression for single-image human shape
  reconstruction.
\newblock In {\em Proceedings of the IEEE/CVF Conference on Computer Vision and
  Pattern Recognition (CVPR)}, June 2019.

\bibitem{ref6_ProHMR}
Nikos Kolotouros, Georgios Pavlakos, Dinesh Jayaraman, and Kostas Daniilidis.
\newblock Probabilistic modeling for human mesh recovery.
\newblock In {\em ICCV}, 2021.

\bibitem{ref_appearance}
Jogendra~Nath Kundu, Mugalodi Rakesh, Varun Jampani, Rahul~M Venkatesh, and
  R.~Venkatesh Babu.
\newblock Appearance consensus driven self-supervised human mesh recovery.
\newblock In {\em Proceedings of the European Conference on Computer Vision
  (ECCV)}, 2020.

\bibitem{rw_ref2_unite}
Christoph Lassner, Javier Romero, Martin Kiefel, Federica Bogo, Michael~J
  Black, and Peter~V Gehler.
\newblock Unite the people: Closing the loop between 3d and 2d human
  representations.
\newblock In {\em Proceedings of the IEEE conference on computer vision and
  pattern recognition}, pages 6050--6059, 2017.

\bibitem{ref_hybrik}
Jiefeng Li, Chao Xu, Zhicun Chen, Siyuan Bian, Lixin Yang, and Cewu Lu.
\newblock Hybrik: A hybrid analytical-neural inverse kinematics solution for 3d
  human pose and shape estimation.
\newblock In {\em Proceedings of the IEEE/CVF Conference on Computer Vision and
  Pattern Recognition}, pages 3383--3393, 2021.

\bibitem{ref_metro}
Kevin Lin, Lijuan Wang, and Zicheng Liu.
\newblock End-to-end human pose and mesh reconstruction with transformers.
\newblock In {\em CVPR}, 2021.

\bibitem{ref_meshgraphormer}
Kevin Lin, Lijuan Wang, and Zicheng Liu.
\newblock Mesh graphormer.
\newblock In {\em ICCV}, 2021.

\bibitem{ref_COCO}
Tsung-Yi Lin, Michael Maire, Serge Belongie, James Hays, Pietro Perona, Deva
  Ramanan, Piotr Dollar, and Larry Zitnick.
\newblock Microsoft coco: Common objects in context.
\newblock In {\em ECCV}. European Conference on Computer Vision, September
  2014.

\bibitem{ref_liu2020attention}
Ruixu Liu, Ju Shen, He Wang, Chen Chen, Sen-ching Cheung, and Vijayan Asari.
\newblock Attention mechanism exploits temporal contexts: Real-time 3d human
  pose reconstruction.
\newblock In {\em Proceedings of the IEEE/CVF Conference on Computer Vision and
  Pattern Recognition}, pages 5064--5073, 2020.

\bibitem{rw_ref15_Soft_Rasterizer}
Shichen Liu, Tianye Li, Weikai Chen, and Hao Li.
\newblock Soft rasterizer: A differentiable renderer for image-based 3d
  reasoning.
\newblock In {\em Proceedings of the IEEE/CVF International Conference on
  Computer Vision}, pages 7708--7717, 2019.

\bibitem{ref1_SMPL}
Matthew Loper, Naureen Mahmood, Javier Romero, Gerard Pons-Moll, and Michael~J.
  Black.
\newblock {SMPL}: A skinned multi-person linear model.
\newblock {\em ACM Trans. Graphics (Proc. SIGGRAPH Asia)}, 34(6):248:1--248:16,
  Oct. 2015.

\bibitem{ref8_MEVA}
Zhengyi Luo, S.~Alireza Golestaneh, and Kris~M. Kitani.
\newblock 3d human motion estimation via motion compression and refinement.
\newblock In {\em Proceedings of the Asian Conference on Computer Vision
  (ACCV)}, November 2020.

\bibitem{ref_mpii3d}
Dushyant Mehta, Helge Rhodin, Dan Casas, Pascal Fua, Oleksandr Sotnychenko,
  Weipeng Xu, and Christian Theobalt.
\newblock Monocular 3d human pose estimation in the wild using improved cnn
  supervision.
\newblock In {\em 3D Vision (3DV), 2017 Fifth International Conference on}.
  IEEE, 2017.

\bibitem{rw_ref13_OccupancyNetwork}
Lars Mescheder, Michael Oechsle, Michael Niemeyer, Sebastian Nowozin, and
  Andreas Geiger.
\newblock Occupancy networks: Learning 3d reconstruction in function space.
\newblock In {\em Proceedings of the IEEE/CVF Conference on Computer Vision and
  Pattern Recognition}, pages 4460--4470, 2019.

\bibitem{ref17_NeRF}
Ben Mildenhall, Pratul~P Srinivasan, Matthew Tancik, Jonathan~T Barron, Ravi
  Ramamoorthi, and Ren Ng.
\newblock Nerf: Representing scenes as neural radiance fields for view
  synthesis.
\newblock In {\em European conference on computer vision}, pages 405--421.
  Springer, 2020.

\bibitem{rw_ref6_I2l}
Gyeongsik Moon and Kyoung~Mu Lee.
\newblock I2l-meshnet: Image-to-lixel prediction network for accurate 3d human
  pose and mesh estimation from a single rgb image.
\newblock In {\em Computer Vision--ECCV 2020: 16th European Conference,
  Glasgow, UK, August 23--28, 2020, Proceedings, Part VII 16}, pages 752--768.
  Springer, 2020.

\bibitem{ref16_GIR}
Michael Niemeyer and Andreas Geiger.
\newblock Giraffe: Representing scenes as compositional generative neural
  feature fields.
\newblock In {\em Proceedings of the IEEE/CVF Conference on Computer Vision and
  Pattern Recognition (CVPR)}, pages 11453--11464, June 2021.

\bibitem{ref15_NeuralBodyFit}
Mohamed Omran, Christoph Lassner, Gerard Pons-Moll, Peter~V. Gehler, and Bernt
  Schiele.
\newblock Neural body fitting: Unifying deep learning and model-based human
  pose and shape estimation.
\newblock Verona, Italy, 2018.

\bibitem{ref_smplify-x}
Georgios Pavlakos, Vasileios Choutas, Nima Ghorbani, Timo Bolkart, Ahmed A.~A.
  Osman, Dimitrios Tzionas, and Michael~J. Black.
\newblock Expressive body capture: 3d hands, face, and body from a single
  image.
\newblock In {\em Proceedings IEEE Conf. on Computer Vision and Pattern
  Recognition (CVPR)}, 2019.

\bibitem{ref_texturepose}
Georgios Pavlakos, Nikos Kolotouros, and Kostas Daniilidis.
\newblock Texturepose: Supervising human mesh estimation with texture
  consistency.
\newblock In {\em ICCV}, 2019.

\bibitem{ref_coarse2fine}
Georgios Pavlakos, Xiaowei Zhou, Konstantinos~G. Derpanis, and Kostas
  Daniilidis.
\newblock Coarse-to-fine volumetric prediction for single-image 3d human pose.
\newblock In {\em Proceedings of the IEEE Conference on Computer Vision and
  Pattern Recognition (CVPR)}, July 2017.

\bibitem{ref13_LearningToEstim}
Georgios Pavlakos, Luyang Zhu, Xiaowei Zhou, and Kostas Daniilidis.
\newblock Learning to estimate 3d human pose and shape from a single color
  image.
\newblock In {\em Proceedings of the IEEE Conference on Computer Vision and
  Pattern Recognition (CVPR)}, June 2018.

\bibitem{ref_pavllo20193d}
Dario Pavllo, Christoph Feichtenhofer, David Grangier, and Michael Auli.
\newblock 3d human pose estimation in video with temporal convolutions and
  semi-supervised training.
\newblock In {\em Proceedings of the IEEE/CVF Conference on Computer Vision and
  Pattern Recognition}, pages 7753--7762, 2019.

\bibitem{ref_delving}
Yu Rong, Ziwei Liu, Cheng Li, Kaidi Cao, and Chen~Change Loy.
\newblock Delving deep into hybrid annotations for 3d human recovery in the
  wild.
\newblock In {\em The IEEE International Conference on Computer Vision (ICCV)},
  October 2019.

\bibitem{rw_ref11_GRAF}
Katja Schwarz, Yiyi Liao, Michael Niemeyer, and Andreas Geiger.
\newblock Graf: Generative radiance fields for 3d-aware image synthesis.
\newblock In {\em Advances in Neural Information Processing Systems (NeurIPS)},
  2020.

\bibitem{ref_hrnet}
Ke Sun, Bin Xiao, Dong Liu, and Jingdong Wang.
\newblock Deep high-resolution representation learning for human pose
  estimation.
\newblock In {\em CVPR}, 2019.

\bibitem{ref_ROMP}
Yu Sun, Qian Bao, Wu Liu, Yili Fu, Black Michael~J., and Tao Mei.
\newblock Monocular, one-stage, regression of multiple 3d people.
\newblock In {\em ICCV}, 2021.

\bibitem{ref_dsd}
Yu Sun, Yun Ye, Wu Liu, Wenpeng Gao, Yili Fu, and Tao Mei.
\newblock Human mesh recovery from monocular images via a skeleton-disentangled
  representation.
\newblock In {\em Proceedings of the IEEE/CVF International Conference on
  Computer Vision (ICCV)}, October 2019.

\bibitem{ref18_Fourier}
Matthew Tancik, Pratul~P. Srinivasan, Ben Mildenhall, Sara Fridovich-Keil,
  Nithin Raghavan, Utkarsh Singhal, Ravi Ramamoorthi, Jonathan~T. Barron, and
  Ren Ng.
\newblock Fourier features let networks learn high frequency functions in low
  dimensional domains.
\newblock {\em NeurIPS}, 2020.

\bibitem{rw_ref9_Self}
Hsiao-Yu Tung, Hsiao-Wei Tung, Ersin Yumer, and Katerina Fragkiadaki.
\newblock Self-supervised learning of motion capture.
\newblock In I. Guyon, U.~V. Luxburg, S. Bengio, H. Wallach, R. Fergus, S.
  Vishwanathan, and R. Garnett, editors, {\em Advances in Neural Information
  Processing Systems 30}, pages 5236--5246. Curran Associates, Inc., 2017.

\bibitem{rw_ref8_BodyNet}
Gul Varol, Duygu Ceylan, Bryan Russell, Jimei Yang, Ersin Yumer, Ivan Laptev,
  and Cordelia Schmid.
\newblock Bodynet: Volumetric inference of 3d human body shapes.
\newblock In {\em Proceedings of the European Conference on Computer Vision
  (ECCV)}, pages 20--36, 2018.

\bibitem{ref_3dpw}
Timo von Marcard, Roberto Henschel, Michael~J. Black, Bodo Rosenhahn, and
  Gerard Pons-Moll.
\newblock Recovering accurate 3d human pose in the wild using imus and a moving
  camera.
\newblock In {\em Proceedings of the European Conference on Computer Vision
  (ECCV)}, September 2018.

\bibitem{ref_mpsnet}
Wen-Li Wei, Jen-Chun Lin, Tyng-Luh Liu, and Hong-Yuan~Mark Liao.
\newblock Capturing humans in motion: Temporal-attentive 3d human pose and
  shape estimation from monocular video.
\newblock In {\em The IEEE Conference on Computer Vision and Pattern
  Recognition (CVPR)}, June 2022.

\bibitem{ref_CBAM}
Sanghyun Woo, Jongchan Park, Joon-Young Lee, and In~So Kweon.
\newblock Cbam: Convolutional block attention module.
\newblock In {\em Proceedings of the European Conference on Computer Vision
  (ECCV)}, September 2018.

\bibitem{ref_GHUM}
Hongyi Xu, Eduard~Gabriel Bazavan, Andrei Zanfir, William~T Freeman, Rahul
  Sukthankar, and Cristian Sminchisescu.
\newblock Ghum \& ghuml: Generative 3d human shape and articulated pose models.
\newblock In {\em Proceedings of the IEEE/CVF Conference on Computer Vision and
  Pattern Recognition}, 2020.

\bibitem{rw_ref10_Denserac}
Yuanlu Xu, Song-Chun Zhu, and Tony Tung.
\newblock Denserac: Joint 3d pose and shape estimation by dense
  render-and-compare.
\newblock In {\em Proceedings of the IEEE/CVF International Conference on
  Computer Vision}, pages 7760--7770, 2019.

\bibitem{ref_PTN}
Xinchen Yan, Jimei Yang, Ersin Yumer, Yijie Guo, and Honglak Lee.
\newblock Perspective transformer nets: Learning single-view 3d object
  reconstruction without 3d supervision.
\newblock In D. Lee, M. Sugiyama, U. Luxburg, I. Guyon, and R. Garnett,
  editors, {\em Advances in Neural Information Processing Systems}, volume~29.
  Curran Associates, Inc., 2016.

\bibitem{ref_weakly}
Andrei Zanfir, Eduard~Gabriel Bazavan, Hongyi Xu, Bill Freeman, Rahul
  Sukthankar, and Cristian Sminchisescu.
\newblock Weakly supervised 3d human pose and shape reconstruction with
  normalizing flows.
\newblock In {\em European Conference on Computer Vision (ECCV)}, pages
  465--481, 2020.

\bibitem{rw_ref5_monocular}
Andrei Zanfir, Elisabeta Marinoiu, and Cristian Sminchisescu.
\newblock Monocular 3d pose and shape estimation of multiple people in natural
  scenes-the importance of multiple scene constraints.
\newblock In {\em Proceedings of the IEEE Conference on Computer Vision and
  Pattern Recognition}, pages 2148--2157, 2018.

\bibitem{ref_pymaf}
Hongwen Zhang, Yating Tian, Xinchi Zhou, Wanli Ouyang, Yebin Liu, Limin Wang,
  and Zhenan Sun.
\newblock Pymaf: 3d human pose and shape regression with pyramidal mesh
  alignment feedback loop.
\newblock In {\em Proceedings of the IEEE International Conference on Computer
  Vision}, 2021.

\bibitem{ref_ochuman}
Tianshu Zhang, Buzhen Huang, and Yangang Wang.
\newblock Object-occluded human shape and pose estimation from a single color
  image.
\newblock In {\em Proceedings of the IEEE/CVF Conference on Computer Vision and
  Pattern Recognition (CVPR)}, June 2020.

\bibitem{rw_ref4_lightweight}
Yuxiang Zhang, Zhe Li, Liang An, Mengcheng Li, Tao Yu, and Yebin Liu.
\newblock Lightweight multi-person total motion capture using sparse multi-view
  cameras.
\newblock In {\em Proceedings of the IEEE/CVF International Conference on
  Computer Vision}, pages 5560--5569, 2021.

\end{thebibliography}
}

\clearpage
\section{Supplementary material}
    \subsection{Implementation Details}
        \label{sec:supp-implementation-details}
        \paragraph{Implementation of ImpHMR.}
        Following the existing single image pose and shape estimation methods~\cite{ref3_HMR,ref4_SPIN,ref_eft,ref_pare}, we use ResNet-50~\cite{ref_resnet} as the backbone model for our image encoder $g$. Note that, because ImpHMR is trained \emph{end-to-end} manner, the weights of the image encoder $g$ are \emph{also updated} during training. For the feature fields module, foreground attention $\mathcal{A}$ adopts the spatial attention proposed in \cite{ref_CBAM} to focus on human-related features. The neural feature field represented by the MLP network $h$ adopts the object feature field suggested in \cite{ref16_GIR}. For the MLP $h$, the size of the input latent and output feature vector is changed to $2048$. For HMR tasks, since it is not necessary to infer the texture of humans, we condition $h$ with a single latent vector $\mathbf{z}_{\text{fg}}$. For volume rendering, we sample $32$ points on a ray direction $\mathbf{r}$, and the spatial resolution of the rendered 2D feature map is fixed to $4\times4$. In positional encoding, the frequency octave is set to $10$ and $4$ for $\mathbf{x}$ and $\mathbf{r}$, respectively. The regressor $\mathcal{R}$ has the same architecture as the regressor in SPIN~\cite{ref4_SPIN} and is initialized with a pre-trained model in~\cite{ref4_SPIN}. The maximum iteration number of $\mathcal{R}$ is set to $3$ as in~\cite{ref3_HMR,ref4_SPIN,ref7_VIBE}. For the geometric guidance branch, deconvolution $\mathcal{D}$ is composed of $5$ layers of transposed convolution along with BN~\cite{ref_bn} and ReLU activation, and outputs a $128\times128$ resolution silhouette. During training, we use Adam~\cite{rw_ref1_ADAM} optimizer with batch size $64$. The $\lambda_{2d}$, $\lambda_{3d}$, $\lambda_{pose}$, $\lambda_{shape}$, and $\lambda_{silh.}$ are set to $300$, $300$, $60$, $0.06$, and $30$, respectively. As in previous works, the learning rate is set to $5e{-}5$, and the network is trained with $50$ epochs.
        
        \vspace{-3mm}
        \paragraph{Implementation of the baseline using PTN.}
        In ablation studies, we use the baseline model, Baseline-\textit{PTN}, in which the feature fields module has been replaced by a voxel-based representation (\ie, PTN~\cite{ref_PTN}) in ImpHMR to verify the efficacy of using implicit representation. To implement Baseline-\textit{PTN}, we use the \textit{Decoder} network proposed in Yan~\etal\cite{ref_PTN} consisting of \textit{Volume Generator} and \textit{Perspective Transformer}. In Baseline-\textit{PTN}, Decoder takes a human-related feature vector $\mathbf{z}_{\text{fg}}$ as input. It generates a feature volume with $4\times4\times4$ resolution, and from the volume generates a projected feature viewed from a specific viewing direction $\phi$ by perspective transform. In order to generate an output of the same size as ImpHMR, we set the channel size of the feature volume to 2048. In addition, since there is no volume density in Baseline-\textit{PTN}, the feature value to be projected is obtained through a max operation when performing perspective projection.

    \vspace{2mm}
    \subsection{Datasets}
        \label{sec:supp-datasets}
        The mixture of 3D (\ie, {\small MPI-INF-3DHP, Human3.6M}) and 2D datasets (\ie, {\small MPII, COCO, LSPET}) is used for training.
        \vspace{-3mm}
        \paragraph{MPI-INF-3DHP~\cite{ref_mpii3d}} is a 3D human pose benchmark mostly taken in indoor environments. The dataset utilizes a markerless motion capture system and contains 3D body joint labels. We use the official training set of 8 subjects and 16 videos per subject for training.
        \vspace{-3mm}
        \paragraph{Human3.6M~\cite{ref_h36m}} is a large-scale indoor 3D human pose dataset containing 3.6 million video frames and corresponding 2D and 3D body joint labels. It consists of 15 action categories and 7 subjects. We use 5 subjects (S1, S5, S6, S7, S8) for training and 2 subjects (S9, S11) for testing. Also, the dataset is subsampled from 50 to 25 frames per second.
        \vspace{-7mm}
        \paragraph{3DPW~\cite{ref_3dpw}} is a 3D human pose dataset obtained with an IMU sensor and RGB camera in an in-the-wild environment. It consists of 60 videos (24 train, 12 validation, 24 test) and 51K frames, annotated with SMPL parameters. We use both for training and evaluation, where noted.
        \vspace{-3mm}
        \paragraph{In-the-wild 2D datasets.} We use MPII~\cite{ref_mpii2d}, COCO~\cite{ref_COCO}, and LSPET~\cite{ref_lspet} datasets for in-the-wild 2D human body keypoint datasets. The MPII, COCO, and LSPET datasets consist of human instances labeled with 2D human body joints in amounts of 14k, 75k, and 7k, respectively. As in PARE~\cite{ref_pare}, we use the pseudo-ground-truth SMPL labels provided by \cite{ref_eft} for each dataset for training.

    \vspace{2mm}
    \subsection{Entanglement Measurement}
        \label{sec:supp-esv}
        To quantitatively evaluate the 3D spatial construction capability of ImpHMR, we define \textbf{E}ntanglement between the \textbf{S}hape and the \textbf{V}iewing direction (in short, ESV). Let $\bm{\beta}_{\phi} = \lbrack\beta_{\phi,1},\cdots, \beta_{\phi,10}\rbrack\in \mathbb{R}^{10}$, where $\beta_{\phi,i}$ denotes $i$-th coefficient in the PCA shape space, be SMPL shape parameters inferred by the model from an arbitrary viewing direction $\phi$. Then, the standard deviation $\sigma_i$ of the $i$-th shape parameters $\beta_{\phi,i}$ for the changing in viewing direction $\phi$ can be obtained as $\sigma_i = \sqrt{\frac{1}{360}\sum_{\phi=0^\circ}^{359^\circ} (\beta_{\phi,i}-\mu_i)^2}$, where $\mu_i$ denotes the mean value of the $i$-th shape parameters calculated by $\mu_i = \frac{1}{360}\sum_{\phi=0^\circ}^{359^\circ} \beta_{\phi,i}$. Finally, we can obtain ESV by averaging the standard deviation $\sigma_i$ of each shape coefficient as $\text{ESV} = \frac{1}{10}\sum_{i=1}^{10} \sigma_i$. The smaller the deviation of the shape parameters inferred for the change in viewing direction, the lower the ESV value is measured. And this indicates that the shape and viewing direction are well disentangled.

    \vspace{2mm}
    \subsection{More Comparison of Inference Speed}
        \label{sec:supp-inference-speed}
        \begin{table}[!t]
        \centering
            \resizebox{0.90\columnwidth}{!}{
            \begin{tabular}{@{}l|r|r|r|r@{}}
            \toprule
                ~~~\small \bf Method ~&~ {\small Res.$1$} ~~&~ {\small Res.$2$} ~~&~ {\small Res.$4$} ~~&~ {\small Res.$6$} ~~~\\
            \midrule
                ~~~METRO~\cite{ref_metro}                   ~~&~ 37.6 ~&~ - ~&~ - ~&~ - ~~\\
                ~~~MeshGraphormer~\cite{ref_meshgraphormer} ~~&~ 18.2 ~&~ - ~&~ - ~&~ - ~~\\
                ~~~HybrIK~\cite{ref_hybrik}                 ~~&~ 23.2 ~&~ - ~&~ - ~&~ - ~~\\
            \midrule
                ~~~ImpHMR (Ours)                            ~~&~ 88.8 ~&~ 88.4 ~&~ 87.1 ~&~ 78.2 ~~\\
            \bottomrule
            \end{tabular}
            }
            \caption{\textbf{Comparison of inference speed.}
            The numbers are in \emph{frames per second} (\textbf{fps}). The Res. denotes the spatial resolution of a 2D feature map in volume rendering for our method. Thanks to efficient spatial representation in feature fields, ImpHMR shows about \emph{$2\sim4$ times faster} fps compared to METRO, MeshGraphormer, and HybrIK.
            }
            \vspace{4mm}
            \label{tbl:supp-inference-speed}
        \end{table}

        Table~\ref{tbl:supp-inference-speed} compares the inference speed (\ie, fps) between ImpHMR and the current best-performing methods~\cite{ref_metro,ref_meshgraphormer,ref_hybrik}. For fair evaluation, we use the lightest backbone model proposed in each paper (R50~\cite{ref_resnet}, HRNet-W64~\cite{ref_hrnet}, and R34~\cite{ref_resnet} for METRO~\cite{ref_metro}, MeshGraphormer~\cite{ref_meshgraphormer}, and HybrIK~\cite{ref_hybrik}, respectively). Also, frames per second (fps) is calculated by averaging the time it took for each model to infer $10000$ times of an input image of $224\times224$ size on RTX 2080Ti GPU. As shown in Tab.~\ref{tbl:supp-inference-speed}, we can notice that ImpHMR has \emph{$2\sim4$ times faster} fps than METRO, MeshGraphormer, and HybrIK, which are latest HMR methods.

    \vspace{2mm}
    \subsection{Qualitative Results}
        \label{sec:supp-qualitative}
        \paragraph{Comparison by rotating the mesh inferred from the canonical viewing direction.} The reason for showing the inferred SMPL mesh viewed from different viewing directions in the Fig. {\color{red}7} (in Sec. {\color{red}4.2}) is to verify that our method has learned well the prior knowledge about human appearance on neural feature fields. In addition, since the MPJPE, PA-MPJPE, and PVE metrics mean reconstruction errors in 3D space, we can expect that our method will show good reconstruction results in other views. For example, as shown in Fig.~\ref{fig:supp_qual}, the mesh inferred from the canonical viewing direction by ImpHMR shows more plausible poses in the side view. Also, the first example in Fig.~\ref{fig:supp_qual} shows the advantage of ImpHMR in the presence of self-occlusion.

        \begin{figure}[!t]
        \centering
            \includegraphics[width=\columnwidth]{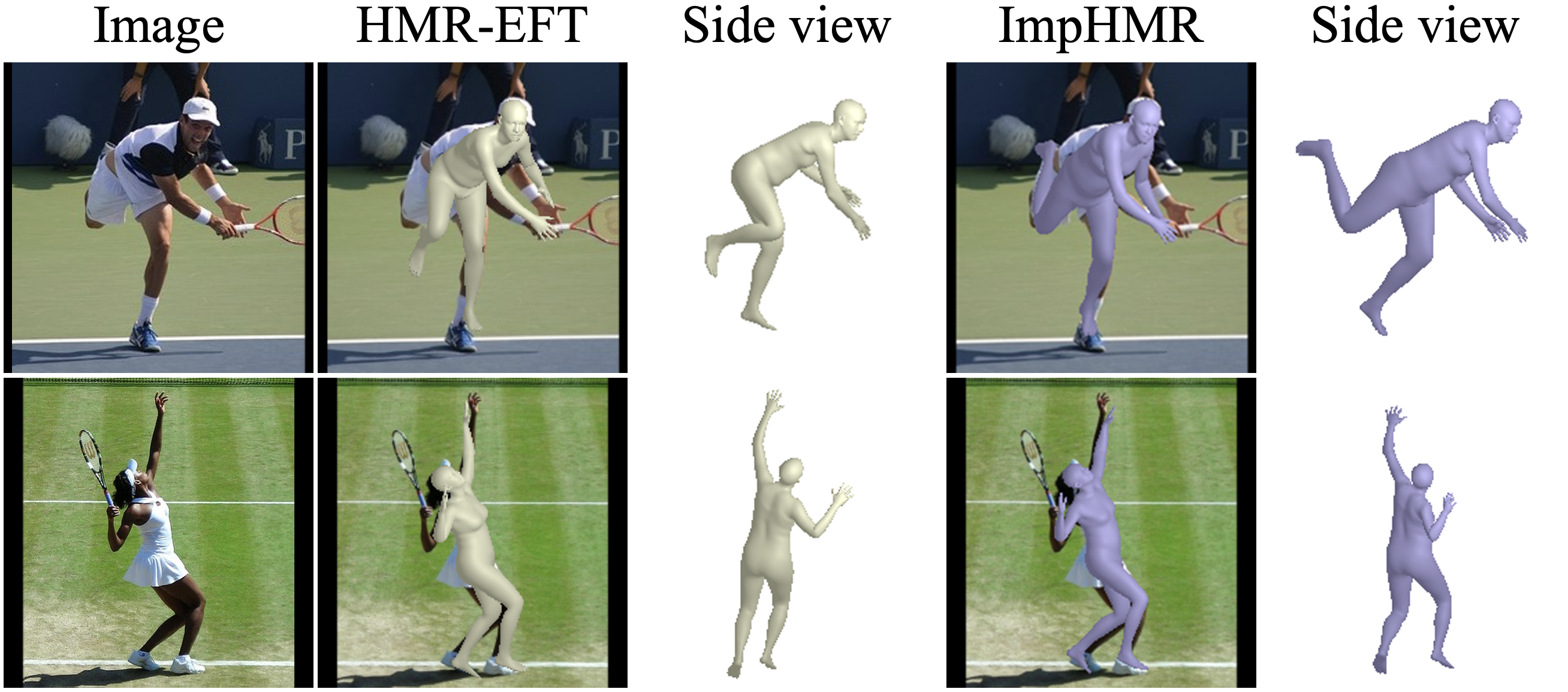}
            \vspace{-4mm}
            \caption{\textbf{Qualitative comparison by rotating the mesh inferred from the canonical viewing direction.}
            For each method, the front and side views of the mesh inferred from the canonical viewing direction are shown in order from left to right.
            }
            \label{fig:supp_qual}
        \end{figure}

    \vspace{2mm}
    \subsection{Failure Cases}
        \label{sec:supp-failure}

        \begin{figure}[!t]
        \centering
            \includegraphics[width=\columnwidth]{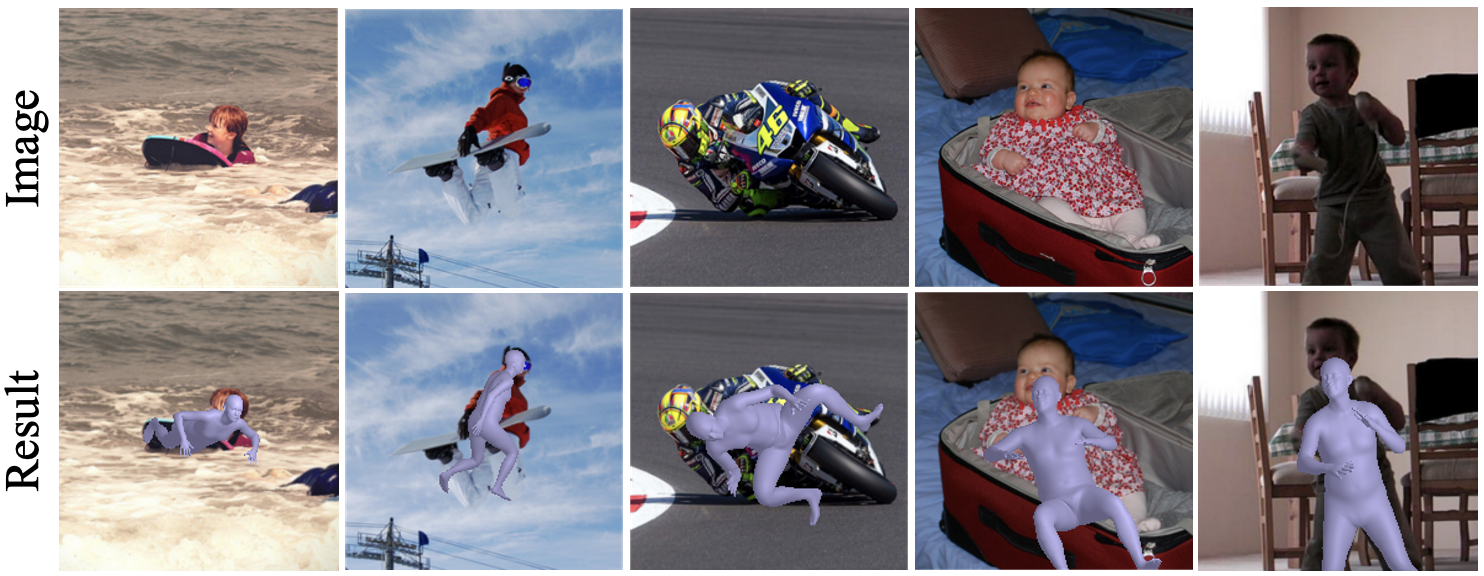}
            \caption{\textbf{Failure cases.}
            Inference results of challenging scenarios (\ie, heavy occlusion, indistinguishable from background, and body shape of children) in which ImpHMR fails to reconstruct pleasing results.
            }
            \label{fig:supp-failure}
        \end{figure}

        Figure~\ref{fig:supp-failure} shows the inference result for challenging scenarios in which ImpHMR fails to reconstruct pleasing results. As can be seen in Fig.~\ref{fig:supp-failure}, ImpHMR fails to infer when most of the human body is occluded, or when the human body is indistinguishable from the background or object, and for the person who has the body shape of children.

\makeatletter
\setlength{\@fptop}{0pt}
\makeatother

\end{document}